\documentclass[aps,superscriptaddress,nofootinbib]{revtex4-2}
\setcitestyle{numbers,open={(},close={)}}

\usepackage[a4paper, left=0.8in, right=0.8in, top=0.7in, bottom=0.7in]{geometry}

\usepackage{amssymb}
\usepackage{amsmath,mathtools}
\usepackage{amsthm}
\usepackage{braket}
\usepackage{float}

\usepackage{lmodern}
\usepackage[T1]{fontenc}

\usepackage{graphicx}%
\graphicspath{{./figures/}}

\usepackage{dcolumn}%
\usepackage{bm}%

\usepackage{hyperref}%
\hypersetup{
	colorlinks=true,
	linkcolor=blue,
	filecolor=blue,
	citecolor=black,      
	urlcolor=cyan,
}

\usepackage{color}
\usepackage{subcaption} %
\usepackage{nicefrac}

\usepackage{orcidlink}

\newcommand{\sign}{\mathrm{Sign}}

\newcommand{\eqnref}[1]{Eq.~\eqref{#1}}
\newcommand{\secref}[1]{Sec.~\ref{#1}}
\newcommand{\figref}[1]{Fig.~\ref{#1}}

\begin{document}

\title{Iterative tensor network transformations for element-wise evaluation of elementary and filtering functions}

\author{Xiao Wang\,\orcidlink{0000-0002-3022-7260}}
\affiliation{%
Clarendon Laboratory, University of Oxford, Parks Road, Oxford OX1 3PU, United Kingdom
}%
\email{xw970921@gmail.com}
\thanks{These authors contributed equally to this work.}

\author{Tomohiro Hashizume\,\orcidlink{0000-0002-7154-5417}}
\affiliation{The Hamburg Centre for Ultrafast Imaging, Luruper Chaussee 149, Hamburg 22761, Germany.}
\affiliation{Institute for Quantum Physics, 
University of Hamburg, Luruper Chaussee 149, Hamburg 22761, Germany}
\email{tomohiro.hashizume@uni-hamburg.de}
\thanks{These authors contributed equally to this work.}

\author{Pia Siegl\,\orcidlink{0000-0003-2249-8121}}
\affiliation{Institute for Quantum Physics, 
University of Hamburg, Luruper Chaussee 149, Hamburg 22761, Germany}
\affiliation{Institute of Software Methods for Product Virtualization, German Aerospace Center (DLR), Nöthnitzer Straße 46b, 01187 Dresden, Germany}

\author{Dieter Jaksch\,\orcidlink{0000-0002-9704-3941}}
\affiliation{%
Clarendon Laboratory, University of Oxford, Parks Road, Oxford OX1 3PU, United Kingdom
}%
\affiliation{The Hamburg Centre for Ultrafast Imaging, Luruper Chaussee 149, Hamburg 22761, Germany.}
\affiliation{Institute for Quantum Physics, 
University of Hamburg, Luruper Chaussee 149, Hamburg 22761, Germany}

\date{\today}

\begin{abstract}
Tensor networks are powerful formats for compressing large-scale data. 
However, their application to general data processing has been limited by the difficulty of performing nonlinear operations. 
Here, we introduce iterative tensor network transformations (ITNTs), a general algorithmic framework for the element-wise evaluation of elementary and nonlinear filtering functions on data encoded as tensor trains (TTs), a class of tensor networks. 
Our approach operates entirely in the compressed domain, enabling efficient computation on exponentially large datasets while maintaining a controlled computational cost.
We demonstrate its power in two key areas: 
(I) evaluating highly nonlinear elementary and filtering functions on a 3D reactive flow field, enabling high-fidelity reaction rate computation and region filtering,
and (II) finding extrema in complex optimization problems, 
such as solving Max-SAT instances on spaces up to $2^{70}$ configurations.
These results establish ITNT as a foundational tool that provides tensor network methods with the capability for general-purpose data science and large-scale optimization.
\end{abstract}


\maketitle

\section{Introduction}
The efficient manipulation of large-scale datasets is a central challenge in modern science,
particularly in the simulation of complex systems and the training of machine learning models.
In quantum many-body physics,
this challenge manifests itself as the \textit{curse of dimensionality},
which has been successfully addressed by tensor network algorithms.
Among these, the tensor train (TT) is a one-dimensional tensor network that represents a data-encoded array of size $N$ as a product of smaller tensors
\cite{mccullochDensitymatrixRenormalizationGroup2007,SCHOLLWOCK201196,orusPracticalIntroductionTensor2014,paeckelTimeevolutionMethodsMatrixproduct2019,verstraeteDensityMatrixRenormalization2023}.
In many cases, the TT structure enables an exponential reduction of the computational and memory requirements to 
$\mathcal{O}(\log N)$.
This powerful data compression has enabled large-scale simulations of quantum systems,
far beyond the reach of methods that operate on the full state vector. Owing to this compression efficiency,
TT algorithms have been extended beyond their quantum origins to classical problems,
including fluid dynamics \cite{lubaschMultigridRenormalization2018,gourianovQuantumInspiredApproach2022,kiffnerTensorNetworkReduced2023,gourianovTensorNetworksEnable2025,vanhülst2025quantuminspiredtensornetworkfractionalstepmethod,hulstQuantumInspiredSimulation2D2026},
plasma physics \cite{huynhQuantumInspiredMachineLearning2023},
and machine learning \cite{hanUnsupervisedGenerativeModeling2018}.
Driven by hardware advancements and algorithmic progress in performing operations,
tensor network simulations are beginning to enable results across diverse physical systems far beyond the regime
where simple decompression into an exact numerical vector format is possible 
\cite{chen2025solving,holscherQuantuminspiredFluidSimulation2025,niedermeier2026solving}.

In this regime, 
the most critical physical quantities in frontier science and engineering are often inherently nonlinear, 
necessitating that the corresponding transformations be performed directly within the compressed TT representation.
This requirement, however, runs against a structural limitation of the format. 
A TT supports only a small set of operations with explicit, rank-controlled implementations:
the linear operations, (partial) integrations that take conditional sums of the element, and, as its only nonlinear primitive, 
the element-wise (Hadamard) product under which the bond dimensions of the factors multiply. 
A generic nonlinear function applied element-wise to a compressed array therefore admits no direct TT implementation. 
One route around this limitation is to construct the transformed data by interpolation 
\cite{michailidisTensorTrainMultiplication2024,fernandezLearningTensorNetworks2025,danisTensortrainWENOScheme2025,meng2026recursive}; 
for complex or steep transformations, however, interpolation fails: sharp discontinuities, emergent singularities, 
and rugged underlying landscapes render the data non-interpolatable.
As Jensen's inequality 
\cite{needhamVisualExplanationJensens1993,bullenHandbookMeansTheir2003} 
illustrates, 
loss of fine structure due to compression or interpolation introduces systematic biases 
when highly nonlinear observables are evaluated 
\cite{jarzynskiNonequilibriumEqualityFree1997,ruelJensensInequalityPredicts1999}. 
This serves as a primary bottleneck to high-accuracy tensor network data processing 
in domains reliant on resolving sharp gradients, 
including potential energy mapping of chemical compounds \cite{jumperHighlyAccurateProtein2021}, turbulent flow simulations 
\cite{Brouzet2021-un,gourianovTensorNetworksEnable2025,hulstQuantumInspiredSimulation2D2026}, 
and statistical models encoding combinatorial optimization problems \cite{dobryninEnergyLandscapesCombinatorial2024}.
In principle, the native operations can be composed into convergent schemes, such as~polynomial expansions built from repeated 
element-wise products, or fixed-point iterations for steeper targets, 
to realize arbitrary functions. 
Whether such repeated, truncated compositions remain accurate for the nonlinear transformations relevant to industrial and scientific applications has, however, not yet been investigated.

\begin{figure}[t!]
   \centering
   \includegraphics[width=0.99\textwidth]{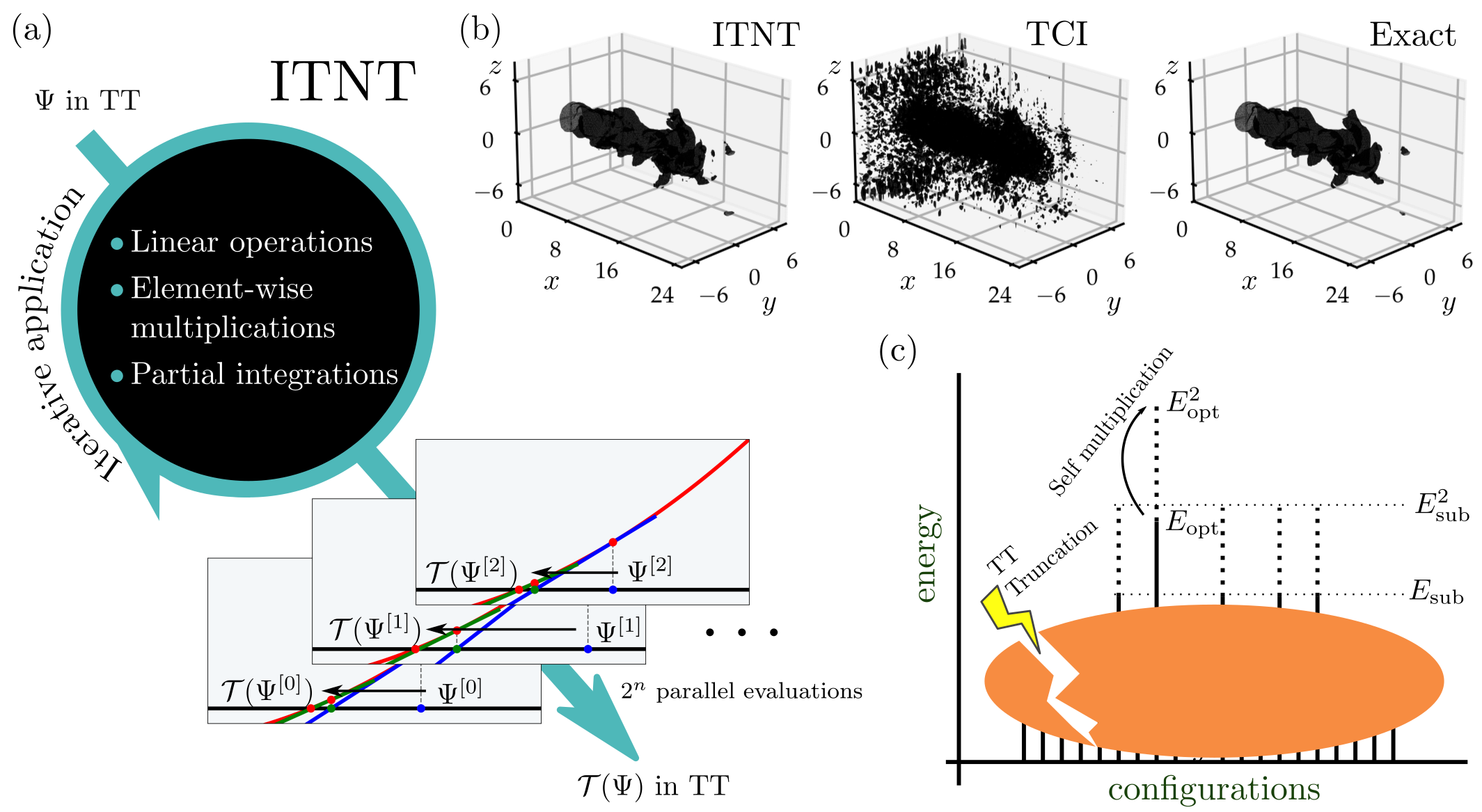}
   \caption{
      {\bf Overview of the results.}
      (a)~Schematic of the Iterative tensor network transformations (ITNTs) methodology.
      The framework enables the evaluation of highly complex nonlinear transformations, $\mathcal{T}$, directly on TT-encoded 
      data $\Psi$ defined on array indices $\bm{j} \in \{0,1,\ldots,N-1=2^{n}-1 \}$. 
      This is achieved by leveraging $2^n$ parallel evaluations of 
      the iteratively converging Newton-Raphson method (bottom-right panels).
      In these panels, the red curve represents the nonlinear function being iteratively solved, 
      while the blue and green lines illustrate the successive tangent-linear approximations during the iterative process, 
      starting from the blue dot.
      This convergence allows the transformation to be applied while maintaining the data in a compressed format. 
      (b) Validation of the framework on a simulated methane/air jet flame \cite{Brouzet2021-un}. 
      The ITNT method is used to compute the highly nonlinear Arrhenius reaction rate ($\exp(-E_c/T)$). 
      The plot shows the iso-rate surface at 10\% of the maximum reaction rate, 
      computed directly from a compressed temperature field $T$ [K] using a normalized reaction energy of $E_c = 17865.2$ [K]. 
      The comparison uses a TT bond dimension of $\chi = 200$, requiring only 0.4%
      The results demonstrate high-fidelity reconstruction, with ITNT achieving enclosed volume relative error approximately
      100 times lower than that of TCI,
      representing a significant improvement in capturing the flame's topological features.
      (c) Schematic of combinatorial optimization utilizing ITNT.
      The framework uses iterative nonlinear operations, specifically self-multiplication, to amplify 
      and sharpen dominant configurations, such as the global maximum ($E_{\mathrm{opt}}$) 
      and sub-optimal peaks ($E_{\mathrm{sub}}$), within the problem-encoded energy landscape.
      Simultaneously, TT truncation (depicted by the lightning bolt) prunes away the rugged bulk spanned by
      the low-energy configurations by discarding the subspace corresponding to smaller singular values.
      This allows the algorithm to isolate optimal solutions while maintaining the data in a compressed TT format. 
   \label{fig:overview}}
\end{figure}

In this article, we address this question by introducing iterative tensor network transformations (ITNTs), 
a general framework for applying arbitrary functions to data encoded as a TT.
ITNTs leverage the iterative application of a small set of efficient TT operations: linear operations, (partial) integrations,
and element-wise multiplications (\figref{fig:overview}~(a)).
This enables the computation of functions ranging from elementary functions 
to highly nonlinear operations, such as filtering, to be applied over $N$ points in parallel, 
performed entirely within the compressed domain in a highly controlled manner. 
To demonstrate the power of ITNTs,
we benchmark their application to large-scale continuous fields and discrete energy landscapes.
For continuous fields, we compute the reaction rate of a 3D reactive flow field 
by applying ITNTs in TT form, giving rise to an order-of-magnitude higher fidelity in comparison to the 
interpolation method (\figref{fig:overview}~(b)). 
For discrete optimization, we find near-optimal configurations for the NP-hard Max-SAT problem with a rugged landscape by filtering out the irrelevant bulk structure via nonlinear operations (\figref{fig:overview}~(c)).
Notably, we identify a solution matching those found by state-of-the-art heuristic algorithms \cite{shen2025free} from a configuration space of $N=2^{70}\approx 10^{21}$ states.
Furthermore, we show that controlled truncation within the ITNT framework enables explicit verification of solution validity, establishing a concrete link between tensor network computational theory and fundamental computational complexity theory.
By enabling the element-wise evaluation of arbitrary functions in a compressed format,
our framework opens an expansive class of problems in large-scale scientific computing, data science, 
and combinatorial optimization to tensor network methods.

\section{Results}
Let $\Psi$ denote an array of length $N$, 
and $\Psi^{[j]}$ ($j\in \{0,\cdots,N-1 \}$) be the $j$\textsuperscript{th} element of this array. 
Here, square brackets are used for superscripts to distinguish them from exponents. 
The TT representation compresses the element-wise numerical representation of $\Psi$ into a product of $n$ local rank-$3$ 
tensors $\Psi[q]$, such that
$\Psi^{[j]} \approx \sum_{\bm{\alpha}} \Psi[0]^{[j_0]}_{\alpha_0,\alpha_1}\Psi[1]^{[j_1]}_{\alpha_1,\alpha_2}\cdots
\Psi[n-1]^{[j_{n-1}]}_{\alpha_{n-1},\alpha_{n}}$. 
In this formulation, $j_{q}\in\{0,1\}$ corresponds to the $q$\textsuperscript{th} bit of the binary 
representation of index $j$, denoted by the sequence $\bm{j} \in \{j_0,j_1,\cdots,j_{n-1}\}$, and $\alpha_r$ corresponds to the bond index of the $r$\textsuperscript{th} bond. 
The summation runs over all the bond indices $\bm{\alpha} = \{\alpha_0,\alpha_1, \cdots,\alpha_{n}\}$,
each of which may have a size up to $\chi$, the maximum bond dimension 
that controls the level of compression (Methods~\ref{sec.svdTrunc}). While a TT can exactly represent any array $\Psi$ given an exponentially large maximum bond dimension $\chi=2^{n/2}$, 
in practice, $\chi$ is kept at a much smaller constant, e.g., $\chi=200$ for \figref{fig:overview}~(b) 
and provides an approximation of $\Psi^{[j]}$.
This value allows us to balance the minimization of compression loss against the reduced memory requirements $\mathcal{O}(n \chi^2)$, 
and the computational scaling $\mathcal{O}(\mathrm{poly}(n)\mathrm{poly}(\chi))$.

To proceed, we introduce the elementary TT operations that constitute the ITNT framework, as 
illustrated in Fig.~\ref{fig:overview}~(a): scalar multiplication, variational addition, 
element-wise products, and partial integration. Specifically, scalar multiplication 
is defined as $\kappa \Psi$, which is implemented by applying the scalar $\kappa$ to a single 
local tensor of the TT. Addition of two TTs $\Psi_1 + \Psi_2$ is a well-established, 
stable routine and requires a computational cost scaling as $\mathcal{O}(n\chi^3)$ \cite{paeckelTimeevolutionMethodsMatrixproduct2019}. 
The element-wise (Hadamard) product, $\Psi_1 \odot \Psi_2$, is realized with a zip-up algorithm using copy tensors 
\cite{biamonteCategoricalTensorNetwork2011}, 
resulting in a scaling of $\mathcal{O}(n\chi^4)$ \cite{stoudenmireMinimallyEntangledTypical2010}
or allowing inflation in the intermediate bond dimension, $\mathcal{O}(n\chi^3)$ \cite{meng2026recursive}.
Finally, the partial integration operator 
$\mathcal{I}(\Psi,q) = \sum_{\bm{j}\backslash \{j_q\} }\sum_{\bm{\alpha}} 
\Psi[0]^{[j_0]}_{\alpha_0,\alpha_1}\Psi[1]^{[j_1]}_{\alpha_1,\alpha_2}\cdots
\Psi[n-1]^{[j_{n-1}]}_{\alpha_{n-1},\alpha_{n}}
$ 
constructs a two-element array by summing all elements for which $j_q=0$ and $j_q=1$, respectively (Methods~\ref{sec.partialintegration}). 

These primitive operations immediately enable 
the element-wise evaluation of integer polynomials and exponentiation directly from their basic definitions.
Consequently, elementary functions such as sine, cosine, hyperbolic sine, and hyperbolic cosine, which are linear combinations of 
(complex) exponentiation, can also be implemented within this framework.
To implement other elementary functions $\mathcal{T}$ \cite{bourchtein2023elementary}, 
particularly those whose inverses are readily computable from the aforementioned primitives, 
we employ the Newton-Raphson method (\figref{fig:overview}~(a), bottom right). 
This involves computing the root $\Psi'$ that satisfies $\mathcal{T}^{-1}(\Psi') - \Psi = 0$, 
where $\mathcal{T}^{-1}(\Psi)$ is the inverse transformation satisfying $\mathcal{T}^{-1}(\mathcal{T}(\Psi))=\Psi$. 
Thus, the elementary functions defined by Bourchtein \textit{et al.}~\cite{bourchtein2023elementary} 
can be fully implemented within the TT framework (for details, see Methods~\ref{sec.elemFunc}). 
To demonstrate the robustness and accuracy of this approach, 
in addition to the methane/air jet flame in \figref{fig:overview}~(b), 
we apply this method to solve the Kidder equation
\cite{iaconoKidderEquation2015,parand2017new,parandGeneralizedLagrangianJacobi2018} in SM~\ref{sec:PDE}.
The Kidder equation is a strongly nonlinear partial differential equation that models flow through porous media with a square-root dependency 
on the field.  
Our numerical results show excellent agreement with state‑of‑the‑art analytical solutions 
on the initial slope of the density field, matching them to six decimal places.

Elementary iterative methods fail for highly nonlinear, non-invertible functions, such as the sign function or
extremum finding, both of which are often required in data processing. 
Here, tools originally developed for homomorphic cryptography \cite{cheonEfficientHomomorphicComparison2020} are required, 
as we will discuss below.

\paragraph{Computing the sign function.} 
Our ITNT sign function algorithm is based on the homomorphic comparison of two encrypted numbers \cite{10.1007/978-3-030-64834-3_8}, which implements comparisons using only addition, scalar multiplication, and element-wise multiplication. 
The TT $\Psi$ is compared to a zero field $\varnothing$ ($\varnothing^{[j]}=0$ for all $j$), yielding an element-wise evaluation of the sign function with the outcome $\sign\{\Psi\}^{[j]}=\sign\{\Psi^{[j]}\}$, 
where $\sign\{x\}=1$ ($x>0$), $\sign\{x\}=0$ for $x=0$, and $\sign\{x\}=-1$ ($x<0$).
The implementation is efficiently achieved via the iteration
$R^{(k+1)} = F\left( R^{(k)} \right)$ with $R^{(0)} = \Psi$ where 
\begin{align}
   F(R^{(k)}) = - \frac{1}{2}\left((R^{(k)})^{\odot 3} - 3R^{(k)}\right) \label{eqn:homomorphF}
\end{align}
and $\Psi^{\odot \theta}$ denotes the $\theta$\textsuperscript{th} element-wise power. 
Convergence of this iteration to \(\sign\{\Psi\} \) requires $ |\Psi^{[j]}| < \sqrt{3}$ ($\forall j$).
Therefore, $\Psi$ might first need to be rescaled to ensure that the iterations remain within the radius of convergence. 
Preconditioning $\Psi$ further may speed up convergence (for details, see  Methods \autoref{sec:accelerationSF}). 
A rough estimate of the normalization constant may be obtained by performing a few iterations of the extremum-finding 
ITNT discussed next.

\paragraph{Extremum-finding ITNT.}
We utilize partial integration to determine, one bit at a time, the configuration that contributes most to the integral.
For a field $\Psi$ with $\Psi^{[j]} \geq 0$, the toal integral obeys 
$\sum_{j_q\in\{0,1\}} \mathcal{I}(\Psi,q)^{[j_q]} = \sum_j \Psi^{[j]}$;
if this sum is dominated by the maximum element $\max \{\Psi^{[j]}\}$, 
then $\mathrm{argmax} \{ \mathcal{I}^{[j_q]}(\Psi,q) \}$
coincides with the $q$\textsuperscript{th} component of the global maximum index $J=\mathrm{argmax}\{ \Psi^{[j]} \}$.
We therefore define the binary fixation (BF) as sequentially fixing the $q$\textsuperscript{th} 
index of a TT to $m_q=\mathrm{argmax}\{\mathcal{I}^{[j_q]}(\Psi,q)\}$, starting from the $0$\textsuperscript{th} tensor. 
BF recovers the full index of the maximum $\bm{m}=(m_0,m_1,\cdots,m_{n-1})=\bm{J}$ after $n$ iterations. 

In practice, a raw field rarely holds such a dominance condition. 
Therefore, as a preprocessing, we propose to raise the field components to a power $2^P$ via self-multiplications 
by performing $P$ steps of self-multiplications. 
This concentrates the field onto its largest elements while preserving the location of the global maximum,
and for sufficiently large $P$, the dominance condition holds for the exact $\Psi^{\odot 2^P}$. 
However, this self-multiplication causes the bond dimension to grow as $\chi^{2^P}$ without bond dimension truncation.
To keep the bond dimension of the output TT manageable,
we use singular value decomposition (SVD) for  truncation \cite{Oseledets2011}. 
SVD, by design, minimizes the $\ell_2$ error, thus the truncation flattens 
and removes the rugged bulk that requires a large bond dimension for its exact representation,
while the landscape near maximal values is preserved due to its large contributions to the global $\ell_2$ norm. 

The $\ell_2$ error, defined as 
$\varepsilon (\Psi , \Psi_{\mathrm{SVD}}) = \| \Psi - \Psi_{\mathrm{SVD}}  \|_2$, 
where $\Psi_{\mathrm{SVD}}$ is the $\Psi$ TT obtained after SVD truncation, 
bounds the $\ell_\infty$ error ($\epsilon_{\infty} = \| \Psi - \Psi_{\mathrm{SVD}}  \|_{\infty}$),  the maximum perturbation added to the elements, from the above.
Thus, the algorithm is guaranteed to converge to a bond dimension of, at worst, $\chi_f$ corresponding 
to the number of indices that give rise to $\max\{\Psi\}$, and BF finds one of these indices as long as 
the accumulated $\ell_2$ error per step at the $p$th self-multiplication step satisfies 
\begin{align}
   \varepsilon (\Psi^{\odot {2^p}} , (\Psi^{\odot {2^p}})_{\mathrm{SVD}}) \leq 
   (\max \{ (\Psi^{\odot 2^p})^{[j]}  \} - \max_{\mathrm{sub}} \{ (\Psi^{\odot 2^p})^{[j]} \})/\sqrt{2}
   \label{eq:normreduction}
\end{align}
for the gap between $\max \{(\Psi^{\odot 2^p})^{[j]} \}$ 
and $\max_{\mathrm{sub}} \{ (\Psi^{\odot 2^p})^{[j]} \}$ the largest element strictly below the maximum 
(cf.~SM~\ref{SM:trunceb} for details). 

When the required number of self-multiplications is intractably large, 
the $\ell_2$ error also becomes large due to the limited bond dimension. 
In such a case, the maximum-finding no longer becomes exact, and one obtains a heuristic, approximate solution, 
which represents the best estimate of $J$ attainable within the available computational resources.
The key is to perform BF in the randomized order, referred to as random binary fixation (RBF),
and remove the identified peak from the landscape (deflation). 
The RBF procedure applies BF iterations to randomly selected sequences of local tensors. 
The subsequent deflation step, $\Psi' \overset{\text{defl}}{\rightarrow} \Psi''$, removes the 
identified configuration from $\Psi'$ such that $\Psi''^{[j]} = \Psi'^{[j]}-\delta_{j,r}\Psi'^{[r]}$
(see Methods \autoref{sec:INT_extremum-finding}), where $r$ is the index found in a given round of RBF. 
After the assigned amount of RBF resource is exhausted, the configuration found that yields the largest value of 
$\Psi$ is accepted as the approximate solution to the global maximum. 

For the rest of this article, we focus on studying the numerical performance of nonlinear functions applied to the most difficult structured data arising from both continuous and discrete problems.
As nonlinear elementary functions, filtering, and extremum-finding represent the most demanding nonlinearities for TT-based numerical methods, they serve as a direct benchmark for the overall robustness and precision of the ITNT framework. 

\paragraph{Application I: Data processing of a 3D reactive flow.}
We first demonstrate the utility of the ITNT framework for large-scale data processing by applying it to high-resolution grid data.
The TT algorithms for extremum-finding and sign-function evaluation can be directly applied to field-processing tasks, 
as illustrated in \figref{fig:overview}~(b). 
Here, we use them to post-process the temperature field of a precomputed methane/air jet flame \cite{Brouzet2021-un}.
This demonstrates how ITNT implements nonlinear functions by computing the Arrhenius reaction rate as an example.

\begin{figure}[t!]
   \centering
   \includegraphics[scale=1.0]{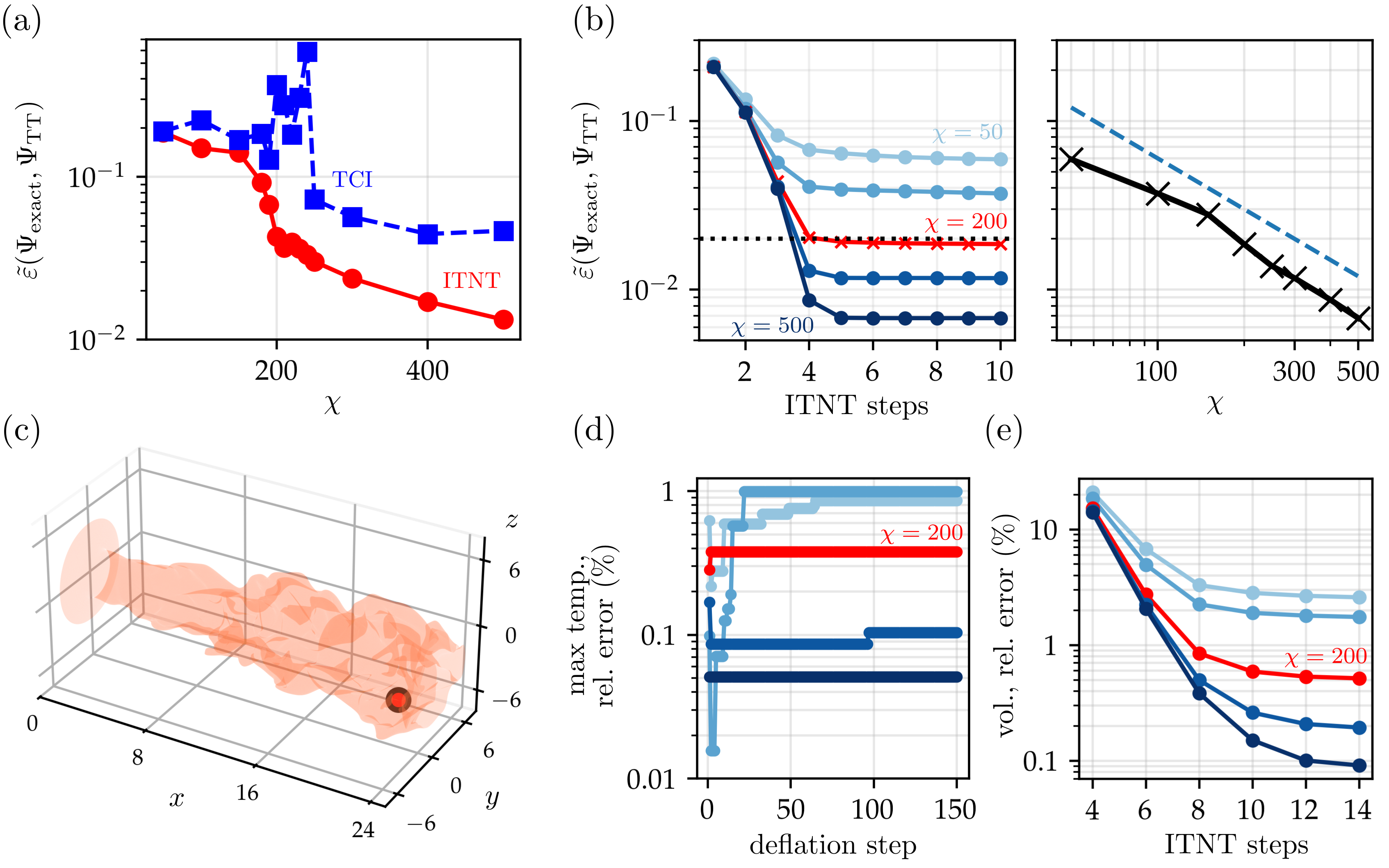}
   \caption{
   \textbf{Nonlinear post processing with ITNTs.}
      (a) Convergence comparison between ITNT and TCI for the Arrhenius equation (\eqnref{eq:arrhenius}) computation.
      While TCI suffers from instability around $\chi=200$, ITNT demonstrates a drop in error in the same region, 
      consistently converging towards the exact result with increasing $\chi$.
      (b) Relative $\ell_2$ error of the $1/T$ field 
      computed exactly ($\Psi_{\mathrm{exact}}$) from the original field
      and by using the ITNT algorithm ($\Psi_{\mathrm{TT}}$) for $\chi=50$, $100$, $200$, $300$, and $500$ 
      (light to dark, left panel),
      and the same error at the 10th iteration for different bond dimensions. 
      In the right panel, the blue dashed line indicates $1/\chi$ decay as a guide to the eye. 
      (c) The position of the maximum temperature is indicated by a red dot and 
      the orange region represents the isothermal surface at $90$\% of the maximum temperature
      ($0.9\max\{T \}$). 
      (d) Relative error of the maximum temperature, $1-\max \{ T_{\mathrm{TT}} \}/\max \{ T_{\mathrm{exact}} \}$,
      found using ITNT for $\chi=50$, $100$, $200$, $300$, and $500$ (light to dark). 
      (e) Relative error of the volume of the region where $T>0.9\max\{ T \}$
      for $\chi=50$, $100$, $200$, $300$, and $500$ (light to dark). 
      Here, the value of $0.9\max\{ T\}$ is computed with the maximum-finding ITNT and the volume is determined
      by constructing the filter with sign-transformation ITNT.
   }
    \label{fig:usecase_jet}
\end{figure}

Formally, the Arrhenius reaction rate is defined as \cite{connors1990chemical},
\begin{align}
   A_r = A_0 \exp(-E_c/T), \label{eq:arrhenius}
\end{align}
where $T$ is the temperature and $E_c=17865.2$ [K] is the activation energy of methane combustion. 
For simplicity, we set the pre-exponential factor $A_0=1$. 
This rate quantifies the chemical kinetics within the flame, which are critical factors in engine design. 

To perform the computation, we encode the temperature field into a TT representation with a maximum bond dimension $\chi$. 
Before encoding, we trim the outer grid points so that each grid dimension is discretized to the nearest power of $2$. 
We then apply the elementary-function ITNT (see Methods~\ref{sec.elemFunc}) in two stages: 
the first involves computing the reciprocal of the temperature field at each point, 
and the second involves exponentiating the result from the previous step.
Throughout these operations, we cap the bond dimension of all intermediate tensors at $\chi$,
where we obtain the TT representation of $A_r$ with a final bond dimension of $\chi$.
Alternatively, for comparison, a Tensor Cross Interpolation (TCI) approach is employed via the \texttt{xfac} library
\cite{fernandezLearningTensorNetworks2025} to map the nonlinear function directly.
This TCI procedure utilizes an iterative pivot-based search, starting from the maximum temperature location to ensure 
a stable initial estimate, refining the tensor cores until the pivot error falls below a tolerance of $\epsilon = 10^{-12}$ 
or the maximum bond dimension of $\chi$ is reached.

As shown in \figref{fig:overview}~(b), 
ITNT provides a good approximation to the true reaction rate features already at $\chi=200$, 
which corresponds to 0.4\% of the original size. 
In contrast, TCI fails to reproduce most of the features and creates artifacts. 
This difference in behavior is further observed in the convergence of the relative $\ell_2$ error 
\begin{align}
   \tilde{\varepsilon}( \Psi_{\mathrm{exact}}, \Psi_{\mathrm{TT}}) 
= \| \Psi_{\mathrm{exact}} - \Psi_{\mathrm{TT}} \|_2/\| \Psi_{\mathrm{exact}} \|_2
   \label{eq:l2error}
\end{align}
between the exact representation $\Psi_{\mathrm{exact}}$ and the TT representation $\Psi_{\mathrm{TT}}$.
As shown in \figref{fig:usecase_jet}~(a), TCI exhibits a region of instability around bond dimension $\chi=200$,
while ITNT shows a steady decrease in error without encountering this instability, 
demonstrating that ITNT succeeds in regimes where the interpolation method fails.

This strong numerical stability and convergence with the bond dimension $\chi$ are further supported by the rapid quadratic convergence enabled by the Newton-Raphson method, as shown in \figref{fig:usecase_jet}~(b) 
for the first step, which computes the reciprocal field ($1/T$ field).
Here, even at $\chi=200$ and after only 5 iterations, 
ITNT converges rapidly and approximates the true $1/T$ field to within a relative error below 2\% (left panel, red).
Furthermore, we observe a $1/\chi$ decay in the error with respect to the bond dimension (right panel).

In \figref{fig:usecase_jet}~(c)-(e), 
we further perform a filtering operation on the temperature field and identify the high-temperature region shown in (c).
To this end, we identify the maximum temperature of the field via the maximum-finding ITNT
and compute the volume of the region with a temperature greater than 90\% of the identified maximum.
Under the naive assumption of underlying smoothness and continuity of the encoded field, 
Bayesian optimization \cite{greenhillBayesianOptimizationAdaptive2020} is known to perform well.
However, as shown in (c), due to chemical reactions and the resulting draft in the temperature field
the maximum temperature occurs in a small, isolated region away 
from the main gas stream and its high-temperature surroundings, 
a typical feature of combustion flows \cite{turns2012introduction}. 
This feature makes standard sampling or learning algorithms inefficient for detecting the maximum.

To find the peak temperature without \textit{a priori} knowledge of the field's exact representation, 
we perform the maximum-finding ITNT, executing five rounds of self-multiplication with a controlled bond dimension $\chi$. 
The subsequent deflation steps (Methods \secref{sec:INT_extremum-finding}) immediately find a candidate value
within 1\% of the true maximum (\figref{fig:usecase_jet}~(d)).
Then, using the sign-transformation ITNT, we compute the volume of the region exceeding 90\% of the found maximum temperature 
for a given $\chi$. 
To evaluate accuracy, we plot the relative error of the volume against the true volume in \figref{fig:usecase_jet}~(e).
Even for a bond dimension as small as $\chi=200$,
a relative error of less than 1\% is achieved after 10 ITNT rounds. 
The continued convergence with increasing $\chi$ demonstrates that this method can perform nonlinear data processing with high accuracy on complex data structures encoded in the TT format (see~SM~\ref{sec:sign_supMat} for further analysis on ITNT transformations acting on continuous data). 

\paragraph{Application II: Solving NP-hard optimization problems}
Beyond continuous fields, we study combinatorial optimization problems as a use case for extremum-finding on TT-encoded 
energy landscapes of discrete configurations.
We focus on the class of combinatorial optimization problems that can be mapped to the energy minimization of 
a classical Ising model with long-range interactions.
Specifically, in the Max-$k$-SAT problem with $n$ variables and $M$ clauses, we want to find a bit string 
$\bm{\sigma}=(\sigma_0,\sigma_1,\ldots,\sigma_{n-1})$ with a corresponding index $\sigma$ that maximizes the following $k$-body Ising-type cost function, 
\begin{equation}
   E^{[\sigma]} = M- \sum_{\mathcal{C}=1}^M \prod_{m=1}^k l_m^\mathcal{C}, \label{eq:maxsatenergy}
\end{equation}
where $l_m^\mathcal{C} \in \{\sigma_{j_{(\mathcal{C},m)}}, 1-\sigma_{j_{(\mathcal{C},m)}}\}$ is chosen based on whether the corresponding literal is negated.
Here, $\mathcal{C}\in \{1,\ldots,M\}$ is the index of the logical clause, and 
$j_{(\mathcal{C},m)}\in[0,n-1]$ denotes the index of the bit appearing in the $m$\textsuperscript{th} term of $\mathcal{C}$. 
Concretely, $l_m^\mathcal{C}$ is defined as a \emph{literal-falsity indicator}: 
we set $l_m^\mathcal{C}=1-\sigma_{j_{(\mathcal{C},m)}}$ for an unnegated literal, 
and set $l_m^\mathcal{C}=\sigma_{j_{(\mathcal{C},m)}}$ for a negated literal. 
With this convention, $\prod_{m=1}^k l_m^\mathcal{C}=1$ if and only if clause $\mathcal{C}$ is violated, 
meaning that $E^{[\sigma]}$ counts the number of satisfied clauses.

Because $E^{[\sigma]}$ counts the number of satisfied clauses, 
it effectively defines the energy landscape of an Ising model with variables $\sigma_j$.
Consequently, the Max-$k$-SAT problem translates into finding the global maximum of this landscape.
We create a TT that encodes the energy landscape of the Max-$3$-SAT problem, which is known to be an NP-hard problem 
\cite{aroracomputationalcomplexity2009},
utilizing the Ising-to-TT encoding (Methods~\autoref{sec:IsingEncoding}). 
We select an instance with $n=70$ variables and $M=700$ clauses from the ``Max-SAT 2016 competition'' dataset \cite{maxsat2016},
specifically, instance \texttt{s3v70c700-1}. 
While an array representation of the energy landscape has $2^{70}$ elements,
the mapped TT consists of $70$ local tensors and has a bond dimension of $\chi=701$.

\begin{figure}[t!]
    \centering
    \includegraphics[width=\textwidth]{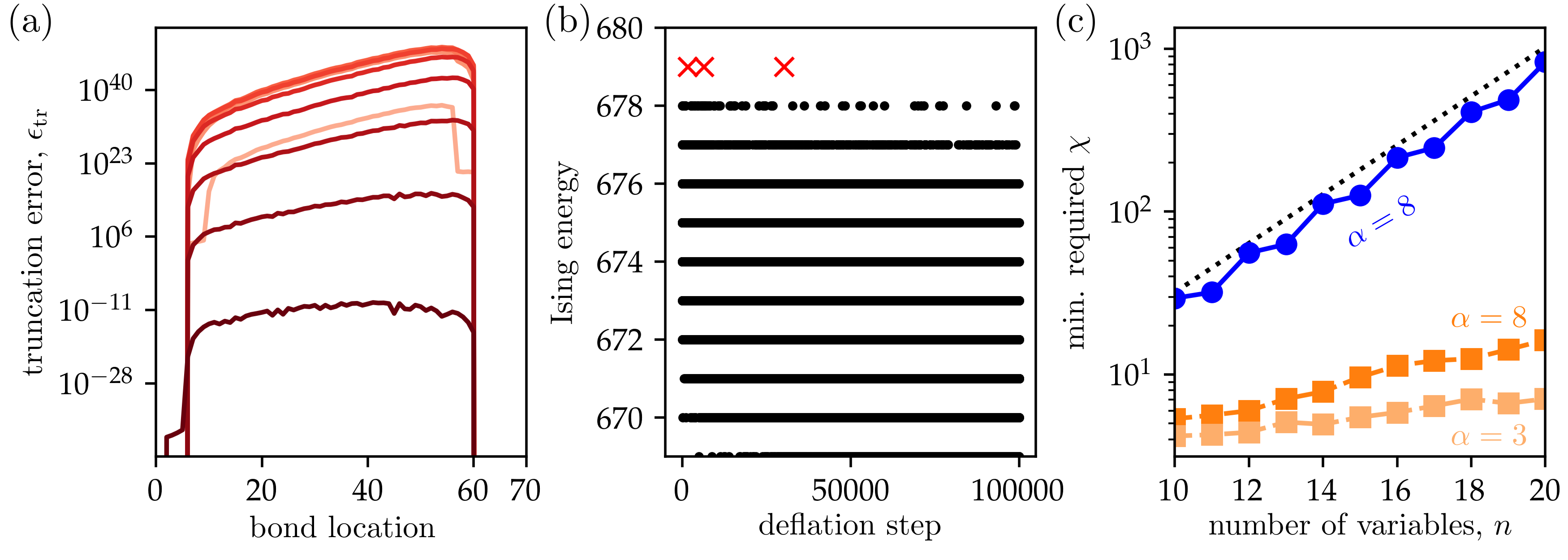}
    \caption{
    \textbf{Solving Max-3-SAT problems with ITNTs.}
      (a) Truncation error at each bond in the zip-up self-multiplication algorithm.
      In order to solve 70 variables 700 clauses Max-3SAT problem, 10 self-multiplications are performed, and for each 
      self-multiplication is computed with the zip-up algorithm with on-site truncation, while fixing $\chi=700$. 
      Each color corresponds to a different self-multiplication iteration (iterations 1 to 10, light to dark)
      (b) Deflation result for the max 3-SAT problem \texttt{s3v70c700-1}. 
      Three configurations with an energy of $E$ with $E=679$ are found after the $i=1655$\textsuperscript{th}, 
      the $i=6442$\textsuperscript{nd}, and the $i=30186$\textsuperscript{th} extremum-finding ITNT steps.
      (c) Scaling of the minimum bond dimension for successful optimization and solution certification.
      Here, we generate 100 random $n$-variable, $M=\alpha n$-clause Max-3SAT problems for $n=10$ to $20$, 
      at clause densities $\alpha=8$ and $\alpha=3$. 
      Orange dashed lines show the minimum bond dimension required for the maximum search to return the correct optimum after
      $P=\left\lceil\log_2(n M) \right\rceil + 1$ self-multiplication steps. 
      The blue line shows the minimum bond dimension required for the truncation budget in Method~\ref{method:truncbudget} 
      to be obeyed after $P$ steps. 
      The required bond dimension grows with $n$, saturating the bond dimension $\sim 2^{n/2}$ (black dotted line). 
    }
    \label{fig:usecase_maxsat}
\end{figure}

To find the maximum number of satisfied clauses and their truth values, 
the landscape TT is first self-multiplied ten times with the bond dimension fixed at $\chi=701$. 
Subsequently, the extremum-locating ITNT (RBF iterations) is performed $10^5$ times. 
During these RBF iterations, the maximum bond dimension is set to $\chi_{\max}=1000$; 
once the TT exceeds this threshold, it triggers a fallback procedure, 
truncating the bond dimension back to $900$.
In \figref{fig:usecase_maxsat}~(a), we plot the truncation error. 
Starting from the first iteration of self-multiplication (lightest), aggressive truncation of the bond dimension causes the truncation error $\epsilon_{\mathrm{tr}} = \sum_{\chi<i} \lambda_i^2$, 
where the singular values $\lambda_i$ are sorted in descending order, to reach $10^{32}$. 
Meanwhile, in the later stages of the self-multiplication iterations, the truncation error gradually decreases, 
showing that the encoded energy landscape converges towards a few dominant configurations (compared to the full $2^{70}$).

Through these iterative steps, we identify three configurations with the extremal value 
$E=679$ (\figref{fig:usecase_maxsat}~(b)). 
These match the best-known solution for this instance and are separated by Hamming distances of $1$, $12$, and $13$:
\begin{center}
\verb|1001001000110000011001111011010101001000000111010010000000110110010101|
\verb|1001001000110000011001111011010101001010000111010010000000110110010101|
\verb|1001001110110000010001111011000101011000011001110010000000111100010101|
\end{center}
Importantly, ITNT identifies these solutions while probing only a tiny fraction of the full configuration space. 
Instead of exhaustively evaluating all $2^{70}\approx 10^{21}$ elements, 
the algorithm visits only $\mathcal{O}(10^4)$ peaks in the landscape. 
The slightly suboptimal extremum (the next-best solution with $E=678$) is identified early in the process after 11 steps, 
providing a high-quality, approximate solution.
This suggests an early-stopping mode for the algorithm; if a near-optimal solution is sufficient, 
the number of deflation steps can be significantly reduced. 
This example illustrates that ITNT has the capability to reach state-of-the-art solutions and is effective in solving discrete NP-hard optimization problems directly within the TT framework.

Furthermore, a fundamental distinction between the ITNT approach and traditional heuristic methods 
is its inherently global search mechanism. 
Unlike traditional solvers, which are sequential and stochastic, 
the algebraic framework aggregates information across the entire ensemble of configurations, 
sidestepping the limitations of incremental, local spin flips. 
By comparing partial sums over an exponential number of configurations in a single operation, 
the RBF procedure identifies global extrema with guaranteed convergence in exact arithmetic, 
given a sufficient number of self-multiplication steps. 
This global comparison is performed efficiently within the TT manifold, offering an exponential advantage over brute-force numerics.

Beyond merely finding a solution, ITNT offers a method to verify whether an identified solution is the true maximum. 
Under truncation, the convergence guarantee above survives only if the error injected at each step remains within a budget set by the margin separating the optimum from the rest of the landscape. 
The certification of the global maximum is done by monitoring the truncation error injected at each self-multiplication step
by assuring the dominance of the true maximum after $P$ self-multiplication steps (Methods~\ref{method:truncbudget}).
However, the unconditional success of such a verification method would imply that 
NP-hard problems can be solved in polynomial time, which would contradict the P$\neq$NP conjecture 
\cite{aroracomputationalcomplexity2009}. 
In \figref{fig:usecase_maxsat}~(c), 
we show the minimum bond dimension required for the truncation-error budget to be satisfied (blue)
and that required for the BF maximum search to succeed throughout $P=\lceil \log_2(n M) \rceil + 1$ steps for 
Max-$3$-SAT problems with $M=\alpha n$ clauses for different values of the hardness parameter $\alpha$. 
As expected from the P$\neq$NP conjecture, 
the minimum bond dimension for certification grows exponentially with the number of variables, saturating the maximum bond dimension bound (black dotted) $2^{n/2}$. 

Finally, we observe that minimum bond dimensions required for successful BF are far smaller than the certification threshold,
which explains why the $n=70$ Max-3-SAT problem could be solved successfully despite the certification threshold being out of reach.
These two results presented in \figref{fig:usecase_maxsat}~(a) and (c) hint at a global optimization that requires far fewer computational resources than brute-force search with a cost that is, for all practical purposes, polynomial in $n$. 
The resource advantage applies to finding solutions, 
while verification of optimality may still require exponential resources.
The presented ITNT-based methods thus shed light on a new computational paradigm offering a novel approach to combinatorial optimization.

\section{Discussion \label{sec:discussion}}
In this article,
we have established iterative tensor network transformations (ITNTs) 
as a general framework for performing element-wise computations on data compressed as a tensor train (TT).
This work fundamentally broadens the scope of tensor network applications,
introducing a method to evaluate arbitrary nonlinear functions entirely within the compressed domain.
Beyond field processing, ITNT introduces a novel optimization paradigm 
distinct from both (quantum) annealing and conventional local heuristic algorithms. 
By comparing partial sums over exponentially large configuration subspaces at each step, 
the extremum-finding algorithm performs a truly global search for optimization tasks where exact and heuristic methods 
often struggle with rugged, multi-modal energy landscapes. 
This demonstrates that ITNT provides high-quality approximate solutions to NP-hard problems within polynomial time under 
a controlled maximum bond dimension, while remaining informed by the solution landscape as a whole.

The performance and ultimate limits of ITNTs represent a rich area for future investigation.
While the zip-up method employed in this article is robust,
alternative approaches may offer superior scaling in different regimes of system size and bond dimension 
\cite{Michailidis2025,meng2026recursive,sun2026stochastic}.
Finally, the convergence of iterative routines could be significantly accelerated through tailored preprocessing steps,
such as initial normalization or the application of a Fourier transform to the TT \cite{chenQuantumFourierTransform2023}.
We anticipate that combining these algorithmic refinements with high-precision arithmetic and GPU acceleration 
\cite{holscherQuantuminspiredFluidSimulation2025,Hauck2025} 
will enable the application of ITNTs to problems on a much larger scale.

This work opens a pathway for applying tensor network methods to a broader class of computationally intensive optimization problems,
including those mappable to Ising Hamiltonians, such as prime factorization \cite{jiangQuantumAnnealingPrime2018}, genetic haplotype reconstruction \cite{zhang2026qhap}, and 
large-scale financial simulations \cite{mansiniHeuristicAlgorithmsPortfolio1999}. 
One long-term direction is to investigate the mapping of ITNT algorithms onto quantum hardware,
where TT-based methods are already showing significant promise \cite{malzPreparationMatrixProduct2024,termanovaTensorQuantumProgramming2024,sieglTensorProgrammableQuantumCircuits2025}.
Furthermore, the ability to encode symmetries directly within the TT formalism \cite{mccullochNonAbelianDensityMatrix2002} 
could be integrated with ITNTs to enforce physical or combinatorial constraints in complex optimization tasks, 
further enhancing the power and reach of this computational paradigm.

\newpage
\section{Acknowledgements}
XW appreciates the insightful comments from Pan Zhang and Li You.
DJ and TH acknowledges support by the European Union’s Horizon Programme (HORIZON-CL42021DIGITALEMERGING-02-10) 
Grant Agreement 101080085 QCFD.
DJ and TH are partially funded by the Cluster of Excellence `Advanced Imaging of Matter' of the Deutsche Forschungsgemeinschaft (DFG)--EXC 2056-project ID 390715994.
PS~acknowledges financial support by the DLR-Quantum-Fellowship Program.
DJ acknowledges support by DFG project ``Quantencomputing mit neutralen Atomen'' (JA 1793/1-1, Japan-JST-DFG-ASPIRE 2024) and the Hamburg Quantum Computing Initiative (HQIC) project EFRE. 
The EFRE project is co-financed by ERDF of the European Union and by the ``Fonds of the Hamburg Ministry of Science, Research, Equalities and Districts (BWFGB)''.

\section{Methods}\label{sec:methods}
In this section, we provide additional details on the foundations and applications of ITNTs discussed in the main text.
First, \autoref{sec.svdTrunc} describes the SVD-based truncation procedure and sweeping protocol used to maintain low-rank tensor representations.
Second, \autoref{sec.partialintegration} outlines the partial integration formalism for tensor trains.
Third, \autoref{sec.elemFunc} explains how to compute elementary functions with ITNT using the Newton-Raphson method.
Fourth, \autoref{sec:accelerationSF} describes preconditioning techniques that accelerate the convergence of element-wise sign
function evaluation.
Fifth, \autoref{sec:INT_extremum-finding} details the extremum-finding ITNT scheme using self-multiplication, 
binary fixation, and deflation.
Finally, \autoref{sec:IsingEncoding} explains how the energy landscape of a classical Ising model is encoded in a compressed 
TT representation. 

\subsection{The SVD truncation} \label{sec.svdTrunc}
The origin of compression in the TT representation lies in the singular value decomposition (SVD),
which identifies the optimal low-rank approximation of a tensor by minimizing the $\ell_2$ norm of 
the truncation error \cite{SCHOLLWOCK201196}. 
By retaining only the subspace that corresponds to the most significant singular values,
SVD truncation ensures the highest-fidelity approximation for a fixed bond dimension $\chi$. 
The careful choice of $\chi$ ensures numerical stability and efficient convergence. 
Furthermore, when performing two-site SVD truncation on a TT representing a smooth function, it is crucial to start the SVD sweep from the smallest scale, 
i.e., truncating from the least significant bit, to achieve efficient compression. 
Notably, reversing the sequence of this sweeping process (i.e., fixing the bond dimension but starting the truncation from the 
large-scale side) leads to a significant increase in the truncation error, emphasizing the critical role of the chosen order. 

\subsection{Partial Integration} \label{sec.partialintegration}
The partial integration $\mathcal{I}(\Psi,q)$ gives a two-element array, 
where $\mathcal{I}(\Psi,q)^{[0]}$ is the sum of all elements of $\Psi^{[j]}$ 
with the $q$\textsuperscript{th} bit of index $\bm{j}$ equal to $0$, 
and $\mathcal{I}(\Psi,q)^{[1]}$ is the corresponding sum for the bit equal to $1$. 
Formally, $\mathcal{I}(\Psi,q)$ is written as
\begin{align}
   \mathcal{I}(\Psi,q)^{[j_q]} = 
   \sum_{\bm{j}\backslash\{ j_q\}}
   \sum_{\bm{\alpha}}
   \Psi[0]_{\alpha_{0},\alpha_{1}}^{[j_0]}
   \cdots
   \Psi[q]_{\alpha_{q},\alpha_{q+1}}^{[j_q]}
   \cdots
   \Psi[n-1]_{\alpha_{n-1},\alpha_{n}}^{[j_{n-1}]}
\end{align}
where $\bm{j}\backslash\{j_q\}$ denotes the bit string $\bm{j}$ without the $q$\textsuperscript{th} bit. 

\subsection{ITNT for realizing elementary functions} \label{sec.elemFunc}
We define the elementary functions as those specified by Bourchtein \textit{et al.}~\cite{bourchtein2023elementary},
comprising polynomial, rational, exponential, trigonometric, and irrational functions 
(i.e., non-integer powers and logarithms). 
First, given the TT $f$ representing the discretized function $f(x)$ over $N=2^n$ grid points, 
we construct the TT $R^{(k)}$
which represents the $k$\textsuperscript{th} integer power of $f(x)$, $\left(f(x)\right)^{k}$, 
using the following iterative element-wise product
\begin{align}
   R^{(k+1)} =  f \odot R^{(k)},
\end{align}
where $R^{(1)} = f$, and the operation $h = f \odot g$ on TTs $f$ and $g$, 
denotes an element-wise product, yielding a resulting TT $h$ with elements $h^{[j]} = f^{[j]}g^{[j]}$. 
With this choice of transform, we obtain $R^{(k)}$, which is equivalent to $f(x)^{\odot k}$. 
We assume that the bond dimensions of both $f$ and $R^{(k)}$ are $\chi$, and this element-wise product is realized using the zip-up method 
by transforming one of them into a diagonal operator using a copy tensor
\cite{biamonteCategoricalTensorNetwork2011,stoudenmireMinimallyEntangledTypical2010}. 
Here, the maximum bond dimension is truncated to $\chi$. 
Under this assumption, 
the element-wise product has a time complexity of $\mathcal{O}(n\chi^4)$ and a memory complexity of $\mathcal{O}(\chi^3)$.

To compute the element-wise exponentiation of the field, $f(x) \to e^{f(x)}$, we rely on the definition of exponentiation as a limit 
\begin{align}\label{eq.exp-definition}
   e^{f(x)} = \lim_{l\to\infty} \left( 1+\frac{f(x)}{2^l}\right)^{\odot 2^l},
\end{align}
where $l\in \mathbb{Z}^{+}$. We can construct this iteration in the TT representation by applying the following transformation $l$ times, 
\begin{align}
   R^{(k+1)} = F\left(R^{(k)}\right) = R^{(k)} \odot R^{(k)},
\end{align}
starting from $R^{(0)} = \mathbb{I} + f / 2^{l}$. 
Here, the unit-element TT $\mathbb{I}$ (i.e., representing a vector with all its elements equal to $1$) 
is constructed from the tensor product of rank-1 unit tensors, 
and thus has a bond dimension of $1$.
As $l$ increases, the final TT approaches the exponentiated TT, $\lim_{l\to\infty} R^{(l)} = \exp\left(f\right)$, 
which represents $e^{f(x)}$.
Although the error in \eqnref{eq.exp-definition} becomes negligible when $2^l \gg f_{\text{max}} \equiv \max_x f(x)$, we still need to restrict the number of iteration steps $l$ to avoid accumulating large truncation errors 
from the element-wise products in the ITNT framework.

This exponentiation ITNT enables the evaluation of thermal path integrals for multi-particle systems, 
ranging from discrete spin models to continuous-space chemical dynamics, directly in the TT representation. 
Specifically, we evaluate the sum of $e^{A(x_t)}$ over all configurations $x_t$,
where each configuration represents a distinct imaginary-time-discretized path.
Using similar iterative self-multiplication techniques,
we can also construct the unitary phase transformation $f(x) \to e^{\mathrm{i} f(x)}$ for complex-valued TT representations,
which serves as a foundation for realizing sine, cosine, and arcsine transforms.
While the convergence of the self-multiplication method may become slow for highly oscillatory functions,
the process can be significantly accelerated using the CORDIC algorithm \cite{muller2006elementary}.
The CORDIC framework decomposes complex functional evaluations into a sequence of elementary, pre-computed rotations.
By utilizing only bit shifts (scalar multiplications by negative integer powers of 2), additions, and comparisons
(as discussed in this article), CORDIC avoids the computational overhead of high-order polynomial expansions. 
This allows us to perform the element-wise evaluation of trigonometric and hyperbolic functions within the TT representation 
(cf.~Ref.~\cite{muller2006elementary} for a modern implementation).

We further expand the ITNT framework by incorporating the Newton-Raphson method. 
The method obtains the solution $R$ to the equation $g(R)=0$ iteratively according to 
\begin{equation}\label{eq.N-R_method}
    R^{(k+1)} = R^{(k)} - \frac{g(R^{(k)})}{g'(R^{(k)})}, 
\end{equation}
where \( g'(R) \) is the derivative of $g$ with respect to $R$.
The Newton-Raphson method exhibits quadratic convergence as long as the function $g(R)$ is differentiable at its root.

This Newton-Raphson method can be used to perform an elementary nonlinear transformation $f(x)\to \mathcal{T}(f(x))$ at each point of the initial function $f(x)$. To achieve this for an arbitrary function $R(x)$, we define
\begin{equation}\label{eq.helper_function_N-R}
    g(R(x)) \equiv \mathcal{T}^{-1}(R(x)) - f(x),
\end{equation}
where $\mathcal{T}^{-1}(\cdot)$ denotes the inverse transformation of $\mathcal{T}(\cdot)$. 
Thus, the solution to $g(R(x)) = 0$ is given by $R(x)=\mathcal{T}(f(x))$.
However, this solution $R(x)$ can also be determined as the fixed point of the Newton-Raphson iterative routine in \eqnref{eq.N-R_method}. The iteration is given by
\begin{equation}\label{eq.N-R_method_Transform}
   R^{(k+1)}(x) = R^{(k)}(x) - \frac{\mathcal{T}^{-1}(R^{(k)}(x)) - f(x)}{(\mathcal{T}^{-1})'(R^{(k)}(x))},
\end{equation}
where $(\mathcal{T}^{-1})'$ denotes the derivative of $\mathcal{T}^{-1}$ with respect to $R^{(k)}(x)$.
Thus, we find that our targeted transformation is given by $\mathcal{T}(f(x)) = \lim_{k\to\infty} R^{(k)}(x)$, 
provided that the initial function $R^{(0)}(x)$ is properly chosen to ensure convergence. 

The above Newton-Raphson equation (\eqnref{eq.N-R_method_Transform}) can be simplified 
and then iteratively computed in the TT representation from the known $f(x)$ and a properly chosen $R^{(0)}$. 
We explain this procedure in the following examples, where the target nonlinear transformation $\mathcal{T}(\cdot)$ respectively represents the inverse, square-root, and logarithmic transformations.

We use the Newton-Raphson method to compute the element-wise reciprocal, $R(x)=1/f(x)$, of the function $f(x)$ using ITNT.
To this end, the equation is reformulated as a root-finding procedure for $g(R(x))=(R(x))^{-1}-f(x)=0$, 
defined in \eqnref{eq.helper_function_N-R}. 
The resulting Newton-Raphson iteration becomes $R^{(k+1)}(x)=R^{(k)}(x)(-f(x)R^{(k)}(x)+2)$. 
This can be rewritten in the TT representation, using the element-wise product and the unit-element TT $\mathbb{I}$, as follows
\begin{equation}\label{eq.ITNT-inverse}
   R^{(k+1)}=R^{(k)} \odot \left( - f \odot R^{(k)} + 2 \mathbb{I}  \right).
\end{equation}
Here, we choose $R^{(0)} = f/c$ as an initial guess, where $c$ is a suppression constant that ensures that the iterations converge.

To compute the square root, $R(x)=\sqrt{f(x)}$, of a function $f(x)$ using ITNT, we again follow \eqnref{eq.helper_function_N-R} and define
$g(R)=R^2-f(x)=0$. The corresponding Newton-Raphson iteration becomes $R^{(k+1)}(x)=\frac{1}{2}\left(R^{(k)}(x)+\frac{f(x)}{R^{(k)}(x)}\right)$, which translates into the TT representation as
\begin{equation}
    R^{(k+1)} = \frac{1}{2}\left(R^{(k)}+ f \odot \left(1/R^{(k)}\right) \right).
\end{equation}
This is also known as Heron's method.
These two operations enable the element-wise evaluation of fractional and negative powers of a field. 
Thus, by iteratively applying these operations, we can achieve the element-wise evaluation of arbitrary real powers.

The natural logarithm, $R(x)=\ln f(x)$, can be computed by defining $g(R)=e^R-f(x)$, resulting in the iterative transformation 
$R^{(k+1)}(x) = R^{(k)}(x) -1 + f(x) e^{-R^{(k)}(x)}$, which translates into the TT representation as follows
\begin{equation}
   R^{(k+1)} = R^{(k)} + f \odot \exp\left(-R^{(k)}\right) - \mathbb{I} .
\end{equation}
We start from $R^{(0)} = 0$. This ensures that $\ln(f(x))=R^{(\infty)}(x)$.
We then compute the exponentiated TT, $\exp\left(-R^{(k)}\right)$, using the ITNT algorithm defined in \eqnref{eq.exp-definition}. 
Convergence is guaranteed when $f(x) > 0$ for all $x$.

\subsection{Accelerating element-wise sign function evaluation \label{sec:accelerationSF}}
It was shown for the sign transformation \cite{10.1007/978-3-030-64834-3_8} that applying the transformation 
\begin{align}
   G(R^{(k)})= - \frac{1}{2^{10}}\left(1359 ((R^{(k)})^{\odot 3}) - 2126R^{(k)}\right) \label{eqn:homomorphG}
\end{align}
before iterating with $F$ (\eqnref{eqn:homomorphF}) further accelerates convergence.
In practice, 
applying the auxiliary transformation $G$ twice serves as an optimal preconditioning step, 
allowing the TT $f$ to converge more rapidly to $\sign\{ f \}$. 
The numerical performance of this iterative composition is illustrated in Fig.~\ref{fig:INT_details}~(b). 
Starting from the identity function $f(x)=x$, 
the composition $F^{(3)} \circ G^{(2)}$ rapidly develops a sharp, 
step-like profile near the origin, effectively approximating the sign function within the compressed TT manifold.

As depicted in the figure, 
the transformation maintains high fidelity for input values within the stable domain; 
however, for $|x|$ exceeding the radius of convergence (indicated by the red shaded regions), 
the iterative map exhibits unstable oscillatory behavior and eventual divergence. 
This divergence occurs because the polynomial composition $F \circ G$ is an iterative map with fixed points at 
$\{-1, 0, 1\}$. 
Input values outside the basin of attraction of these points are driven towards infinity, 
necessitating the global rescaling protocol discussed previously to ensure that all elements of the field remain within the convergence regime.

\subsection{Extremum-finding ITNT with a limited number of self-multiplications\label{sec:INT_extremum-finding}}

\begin{figure*}[t!]
    \includegraphics[width=0.99\linewidth]{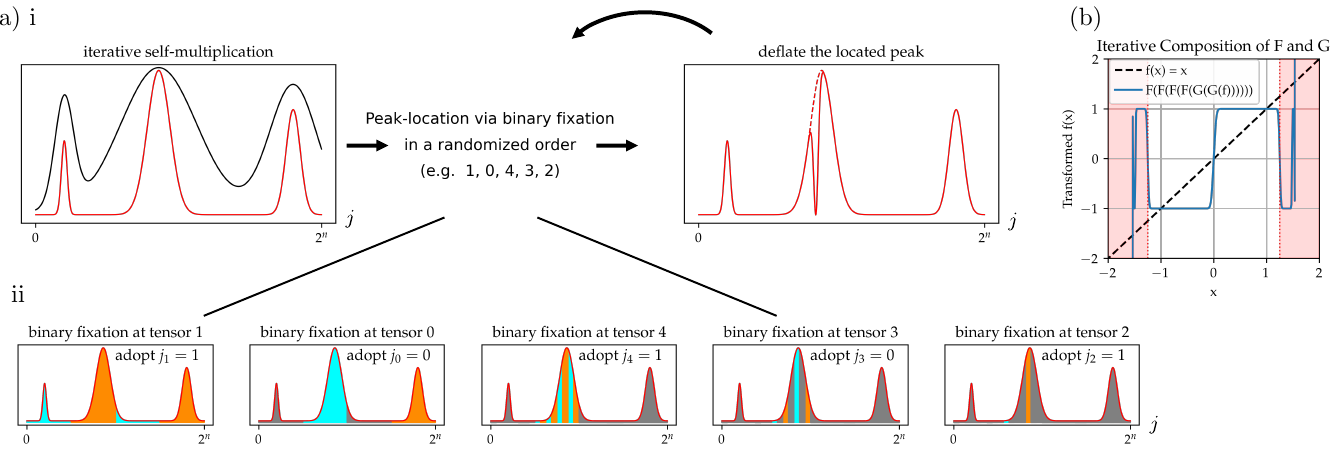}
    \caption{
      \textbf{ITNTs for element-wise nonlinear operations in TT representation.}
      (a) Extrema detection with TT.
      i~Starting from a TT representation of $\Psi$ (black curve), 
      repeated element-wise self-multiplication sharpens the state into a multi-peak profile $\Phi$, 
      where its integral is dominated by extremal configurations (red curve). 
      ii~A peak is then located by a binary fixation procedure.
      This involves integrating out all unfixed sites and choosing the local configuration that maximizes the summed value. 
      Starting from every index not being fixed, in the leftmost panel, 
      a comparison between $j_2=0$ (orange) and $j_2=1$ (cyan) is performed,
      identifying $j_2=1$ as a candidate for the optimal configuration. 
      In the subsequent next panel, $j_1=0$ (orange) and $j_1=1$ (cyan) are compared and $j_1=0$ is selected,
      while the integrated out region from the fixation of $j_2=0$ is indicated in gray.
      Continuing these steps with a random sequence of local tensors identifies the 
      index $I$ of the maximum (orange, right most panel). 
      iii~The deflation process then removes the identified peak to ensure that subsequent applications 
      find the true maximum. 
      (b) Element-wise evaluation of sign function. A highly nonlinear, 
         non-analytic element-wise filter is realized through the composition of iterative applications of the functions
      $F$ (\eqnref{eqn:homomorphF}) and $G$ (\eqnref{eqn:homomorphG}) that produce an element-wise sign
      function $\sign\{f\}^{[j]}=\sign\{f^{[j]}\}$. 
      As an illustrative example, the blue curve shows the output of ITNT for the input field $f(x)=x$ 
      (black dashed line) after applying $G$ twice followed by repeated iterations of $F$. 
      The red shaded region marks the regime of input values where the composed iteration $F(\ldots F(G(G(f))))$ 
      does not converge. 
      The curves are evaluated in the full-array representation solely to visualize the mechanism of the procedure.
      }
    \label{fig:INT_details}
\end{figure*}

Following the main text and \figref{fig:INT_details}~(a), in this subsection, we describe the ITNT algorithm for extrema location in more detail. 
Given an $n$-tensor TT $f$ encoding a real-valued field, 
we amplify large elements by iterative element-wise self-multiplication,
\begin{align}
   R^{(k+1)}&=R^{(k)}\odot R^{(k)} &(R^{(0)}=f).
\end{align}
We then apply normalization after each step,
\begin{equation}
    R^{(k)} \leftarrow \frac{R^{(k)}}{\|R^{(k)}\|} .
\end{equation}
After $k$ iterations, $R^{(k)}$ represents $f^{\odot 2^k}$ up to normalization and typically develops 
a small set of dominant peaks (cf.~\figref{fig:INT_details}~(a)~i.).
Since self-multiplication squares the elements, the peaks appear at indices where $|f^{[j]}|$ is large. 

\paragraph{Peak location with binary fixation}
To extract a candidate index $I$ from a multi-peak TT $R$, we randomly choose a site ordering $(p_0,\dots,p_{n-1})$.
At a binary fixation step $t$, with sites $\{p_0,\dots,p_{t-1}\}$ already fixed, we compare the magnitudes of the two partial 
integrations for $p_t=0$ and $p_t=1$, after all other free tensors have been summed out. 
We then set $p_t$ to the branch with the larger magnitude, and continue this process until all sites are fixed, yielding a candidate index $I$ 
(\figref{fig:INT_details}~(a)ii).

\paragraph{Deflation}
Once a candidate index $I$ yielding a maximum element of $R$ is obtained, we remove the corresponding peak from $R$ by subtracting its contribution (Fig.~\ref{fig:INT_details}~(a)iii)
\begin{equation}
   R \leftarrow R - R^{[I]}\delta_{I},
\end{equation}
where $\delta_{I}$ is a $\chi=1$ TT whose local tensors are given by $\delta[q]^{[i]}_{0,0}=\delta_{i_q,I_q}$. 
Here $\delta_{a,b}$ is a Kronecker delta, defined as $\delta_{a,b}=0$ ($a\neq b$) and $\delta_{a,b}=1$ ($a= b$).
Iterating binary fixation $\to$ deflation produces a ranked list of candidate maxima, 
from which the largest is accepted as the true maximum. 

In principle, choosing a sufficient number of self-multiplications followed by BF 
allows one to identify the true maximum with certainty.
However, the iterative self-multiplication step leads to an exponential growth in the bond dimension.
Thus, intermediate truncations of the bond dimension are necessary but will introduce truncation errors in most cases. 
We find that the accumulation of truncation errors and round-off errors in double-precision arithmetic 
can prevent the ITNT algorithm from reliably locating the global extremum, 
especially if the energy landscape features near-degenerate extrema.

This can be mitigated by randomly traversing the site order across different binary fixation steps
and performing deflation.
The randomized binary fixation sequence at each iteration prevents the algorithm from getting stuck
in local extrema. While a fixed sequence like $(0, \dots, n-1)$ 
tends to yield suboptimal peaks with redundant leading configurations,
the stochastic selection of the fixation sites ensures that the global maximum eventually emerges,
overcoming earlier truncation and round-off errors 
(see SM~\ref{SM:bench-2DNNI} for an application in energy matching in the Ising model).

\subsection{Encoding the energy landscape of Ising Hamiltonians}\label{sec:IsingEncoding}
In this section, we explain how the energy landscape of the classical Ising model can be encoded into a TT. 
We define the classical Ising model on $n$ sites with the Hamiltonian 
\begin{equation}\label{eq.hamiltonian}
   H = \sum_{i=1}^{n}\sum_{j=1}^{i} J_{ij}\sigma_i\sigma_j,
\end{equation}
where $\sigma_i\in \{0,1\}$ are the spin variables. 
Here, we consider Hamiltonians with only two-body interactions; however, the result trivially generalizes to 
Hamiltonians with arbitrary $m$-body interactions. 
For a given configuration $\bm{\sigma}=(\sigma_0,\sigma_1,\ldots,\sigma_{n-1})$, the energy is given by
\begin{equation}\label{eq.landscape}
   E(\bm{\sigma}) = \sum_{i=1}^{n}\sum_{j=1}^{i} J_{ij}\sigma_i\sigma_j. 
\end{equation}
Here, $J_{ij}$ denotes the Ising couplings for $i\neq j$, 
and $J_{ii}$ represents the on-site terms, since $\sigma_i^2=\sigma_i$.

To find the configuration that maximizes $E(\bm{\sigma})$ by performing the extremum-finding ITNT,
we encode the energy landscape into the TT $E$ by setting the values of $E$ at index $\sigma$ to the energy $E(\bm{\sigma})$,
i.e., $E^{[\sigma]}=E(\bm{\sigma})$, where $\sigma$ is the index obtained by interpreting the configuration as a bit string. 
The TT representation of the energy landscape $E(\bm{\sigma})$ is sparse.
The bond dimension $\chi$ of $E$ is bounded by the number of non-zero coupling terms in $J_{ij}$, because we can decompose $E$ as $E = \sum_{i=1}^{n}\sum_{j=1}^{i} E[i,j]$, 
where $E[i,j]$ is a TT encoding the coupling energy corresponding to the term $J_{ij}$ across all configurations. 
Crucially, $E[i,j]$ can be recast into a $\chi=1$ TT as follows, for $i\neq j$, 
\begin{align}
   E[i,j]^{[k]} 
   = J_{ij} \sum_{\bm{\alpha}} &A[0]_{\alpha_0,\alpha_1}^{[k_0]}
   \cdots A[i-1]_{\alpha_{i-1},\alpha_{i}}^{[k_{i-1}]} B[i]_{\alpha_{i},\alpha_{i+1}}^{[k_{i}]}
   A[i+1]_{\alpha_{i+1},\alpha_{i+2}}^{[k_{i+1}]} \cdots  \nonumber\\
   & A[j-1]_{\alpha_{j-1},\alpha_{j}}^{[k_{j-1}]} B[j]_{\alpha_{j},\alpha_{j+1}}^{[k_{j}]}
   A[j+1]_{\alpha_{j+1},\alpha_{j+2}}^{[k_{j+1}]} \cdots A[n-1]_{\alpha_{n-1},\alpha_{n}}^{[k_{n-1}]},
   \label{eq.MPS-encoding-energy-landscape}
\end{align}
where $A[q]$ is the unit tensor with all elements equal to $1$, and $B[q]$ is a $\chi=1$ local tensor 
which takes the form $B[q]^{[k_q]}_{0,0}=k_q$, for a given site $q$. 
To encode the energy landscape of any $m$-body interaction term, $E[i_1,i_2,i_3,\cdots,i_m]$, 
one creates the unit TT $\mathbb{I}$ and
then replaces the local tensors corresponding to the sites that appear in the interaction term with the $B$ tensor. 

According to Ref.~\cite{SCHOLLWOCK201196}, the sum of two TTs yields another TT with a summed bond dimension.
Thus, summing over $E[i,j]$ results in the TT $E$ having a bond dimension 
equivalent to the number of terms in $E(\bm{\sigma})$, including the constant offset. 
Because of this sparsity, we perform ITNTs directly on $E$ to extract various features of the energy landscape: the lowest- and highest-energy configurations, as well as configurations closest to a specified energy value. 
We view this as a global transformation of the energy landscape in its TT representation, which acts simultaneously across all spin configurations. 
This approach differs from tensor network simulations of quantum annealing \cite{tindall-2025}, where the TT encodes the quantum state of the annealer rather than the energy landscape considered here.
\subsection{Truncation error budget for certified optimality of Max-3-SAT solutions}\label{method:truncbudget}
Finally, we derive a truncation-error bound extending \eqnref{eq:normreduction} 
that certifies the maximum found by the maximum-finding ITNT. 
First we derive the sufficient number $P$ of self-multiplication steps for the field $E$ that encodes the energy landscape of the $M$-clause Max-3-SAT problem 
as defined in \eqnref{eq:maxsatenergy}. 
We then define the normalized field $\Psi = E/A$, 
where $A = \max \{ E \} $ is the maximum number of satisfied clauses. 

A value of \(P\) sufficient for BF to retain an optimal branch is obtained by considering the worst case in which every nonoptimal element of $\Psi$ takes its largest possible value, \(1-1/A\).
Let $J$ be the index with $\Psi^{[J]} = 1$, assumed unique for simplicity; degeneracy of the optimum only increases with the margin below. 
The sum of all elements of $\Psi^{\odot 2^P}$ other than $J$ then obeys 
\begin{align}
   \sum_{j \neq J} (\Psi^{\odot 2^P})^{[j]}  \leq 2^{n}(1-1/A)^{2^P} \leq 2^{n}\exp\left(-2^{P}/M\right),
\end{align}
where the right-hand inequality uses $A\leq M$, valid since at most $M$ clauses can be satisfied.
It is therefore sufficient to choose $P$ such that $2^{n}\exp\left(-2^{P}/M\right) <1$,
and hence we obtain $P > \log_2(nM \ln 2)$. 
For the result in \figref{fig:usecase_maxsat}(c) we set $P=\left\lceil\log_2(nM)\right\rceil + 1$, which obeys the bound. 

Given $P$ satisfies the bound, 
we now derive the truncation budget for this dominance condition to be met at the end of $P$ self-multiplication steps.
Let $R^{(p)}$ denote the computed field after $p\leq P$ steps, with $R^{(0)}=\Psi$ and 
\begin{align}
   R^{(p)} = (R^{(p-1)})^{\odot 2} + \tau_p 
\end{align}
where $\tau_p$ is the total error injected by the truncation at step $p$. 

Let $t_p=\varepsilon((R^{(p-1)})^{\odot 2},R^{(p)})=\left\| \tau_p \right\|_2$ be the $\ell_2$ 
norm of the error as defined in \eqnref{eq:normreduction}
and let $\Delta_p=R^{(p)} - \Psi^{\odot 2^p}$ be the error tensor, 
the accumulated deviation from the true representation of the self-multiplied field.

After $P$ iterations, at each step BF compares two partial integrals over the sites not yet fixed 
and keeps the larger. 
Suppose all the previous steps have identified the bits of $J$ correctly.
The difference between the partial integral at the $q$th site with the correct bit 
($j_q=J_q$) and with the incorrect bit ($j_q \neq J_q$) is then bounded from below by
\begin{align}
   \Lambda(\Psi^{\odot 2^P},q)
   \geq 1 - \sum_{i \neq J} (\Psi^{\odot 2^P})^{[i]} \geq 1 - 2^n \exp(-2^{P}/M),
\end{align}
where
$\Lambda(X,q) = \mathcal{I}(X,q)^{j_q = J_q} - \mathcal{I}(X,q)^{j_q \neq J_q}$
for a given field $X$ is the difference between the branches, one with the solution and the other without, and the second inequality holds since $J$ contributes $1$ to the first integral and every remaining configuration contributes non-negatively to the second. 
Invoking the triangular inequality, for $R^{(p)}$ to identify $J$ correctly it therefore suffices that
\begin{align}
   \Lambda(R^{(P)},q) 
   \geq 1 - 2^n \exp(-2^{P}/M) 
   - \left|\Lambda(\Delta_P,q)\right| > 0. \label{ineq:labmdaconst}
\end{align}

From the definition, $\Delta_p$ obeys the recursion
\begin{align}
   \Delta_p = 2\Psi^{\odot 2^{p-1}} \odot \Delta_{p-1} + \Delta^{\odot 2}_{p-1} + \tau_p. 
\end{align}
From this, provided that the initial truncation is $0$, we derive
\begin{align}
   \Delta_P &= 2^{P-p}\Psi^{\odot (2^{P}-2^{p})}\odot \Delta_p 
   + \sum_{m=p+1}^{P} 2^{P-m}\Psi^{\odot (2^{P}-2^{m})}\odot \Delta_{m-1}^{\odot 2} 
   + \sum_{m=p+1}^{P-1} 2^{P-m}\Psi^{\odot (2^{P}-2^{m})}\odot \tau_m
   \nonumber\\
            &= \sum_{m=1}^{P} 2^{P-m}\Psi^{\odot (2^{P}-2^{m})}\odot\Delta_{m-1}^{\odot 2}
            + \sum_{m=1}^{P-1} 2^{P-m}\Psi^{\odot (2^{P}-2^{m})}\odot\tau_m, \label{ineq:Deltap}
\end{align}
which is merely a telescoping error propagation. 

Finally, let $u_p = 2u_{p-1} + u^{2}_{p-1}+t_p$ with $u_0=0$, 
combining the linearity of the partial integration with inequality \eqnref{ineq:Deltap} and applying 
the bounds
\begin{align}
   |\Lambda(\Psi^{\odot (2^{P}-2^{m})}\odot\tau_m,q)| \leq \|\Psi^{\odot (2^{P}-2^{m})}\|_2\|\tau_m \|_2
   \leq \sqrt{2}t_m
\end{align}
and
\begin{align}
   |\Lambda(\Psi^{\odot (2^{P}-2^{m})}\odot\Delta_{m-1}^2,q)| \leq \|\Psi^{\odot (2^{P}-2^{m})}\|_{\infty}\|\Delta_{m-1} \|_2
   \leq u_{m-1}^2,
\end{align}
we obtain 
\begin{align} 
\left| \Lambda\!\left(\Delta_P,q\right) \right| 
\leq \sum_{m=1}^{P-1} 2^{P-m}\sqrt{2}\,t_m + \sum_{m=1}^{P} 2^{P-m}u_{m-1}^2. 
\end{align}
Here, the last truncation is omitted since fusing the final squaring into the BF contraction requires no compression ($t_P=0$).
Together with \eqnref{ineq:labmdaconst}, the optimality of the solution is certified provided
\begin{align}  
   \sum_{m=1}^{P-1} 2^{P-m}\sqrt{2}\,t_m + \sum_{m=1}^{P} 2^{P-m}u_{m-1}^2   < 1-2^n\exp\left(-\frac{2^P}{M}\right).  
\end{align}
This is exactly the bound that the truncation errors $t_m$ must obey over the full $P$ steps for the optimality of the solution to be certified. 

\bibliography{bibliography}

\clearpage

\onecolumngrid
\setcounter{secnumdepth}{2}  %
\pagenumbering{arabic}   %
\setcounter{page}{1}

\renewcommand{\thefigure}{S\arabic{figure}}
\setcounter{figure}{0}

\renewcommand{\thesection}{\Roman{section}}
\renewcommand{\thesubsection}{\Roman{section}.\Alph{subsection}}
\renewcommand{\theequation}{S\arabic{equation}}
\renewcommand{\subsectionautorefname}{Section}

\setcounter{section}{0}
\setcounter{equation}{0}

\makeatletter
\renewcommand\section{\@startsection{section}{1}{0pt}%
  {24pt plus 6pt minus 6pt}%
  {16pt plus 6pt minus 4pt}%
  {\centering\normalfont\normalsize\bfseries}} %
\makeatother

\begingroup
\centering
\large\bfseries Supplemental Material for
Iterative Tensor Network transformations for Element-Wise Evaluation of Elementary and Filtering Functions \par
\vspace{6pt}
\normalfont\normalsize   %
Xiao Wang$^{1}$, Tomohiro Hashizume$^{2,3}$, Pia Siegl$^{3,4}$, Dieter Jaksch$^{1,2,3}$ \par
\vspace{2pt}
\textit{$^{1}$ Clarendon Laboratory, University of Oxford, Parks Road, Oxford OX1 3PU, United Kingdom}\\
\textit{$^{2}$The Hamburg Centre for Ultrafast Imaging, Hamburg 22761, Germany.}\\
\textit{$^{3}$Institute for Quantum Physics, 
University of Hamburg, Luruper Chaussee 149, Hamburg 22761, Germany.}\\
\textit{$^{4}$Institute of Software Methods for Product Virtualization, German Aerospace Center (DLR), Nöthnitzer Straße 46b, 01187 Dresden, Germany}\\
\par
\vspace{8pt}

\justifying
\par
\endgroup
\vspace{12pt}

\section{Application in nonlinear PDE\label{sec:PDE}}
In this section, we demonstrate the application of the Newton-Raphson-based ITNT method in \autoref{sec.elemFunc} to solve nonlinear PDEs in the TT representation, with particular focus on the Kidder equation, which models flow pressure in porous tubes.
\subsection{Using the Newton-Raphson Method to realize the nonlinearity in Kidder equation}\label{sec.kidder}
To solve the Kidder equation, we need to realize the element-wise transformation \( w(x) \to \bigl(1-\kappa w(x)\bigr)^{-\frac{1}{2}} \) in TT representation. 
Following \autoref{sec.elemFunc}, we obtain the solution of \( g(R(x)) = \frac{R(x)^{-2} - 1}{-\kappa} - w(x)  =0 \), 
via the iteration:
\[
R^{(k+1)}(x) = R^{(k)}(x) - \frac{R^{(k)}(x)^{-2} - \bigl(1 - \kappa w(x)\bigr)}{-2 R^{(k)}(x)^{-3}},
\]
Simplifying this expression, we obtain the following iteration in TT representation:
\begin{equation}\label{eq.kappaTransform}
   R^{(k+1)} = \frac{1}{2} R^{(k)} \odot \left( 3 \mathbb{I} - \bigl(1 - \kappa w\bigr) \odot R^{(k)}\odot R^{(k)} \right). 
\end{equation}
Here, $R^{(k)}$ and $w$ are TT encoding the function $R^{(k)}$ and $w(x)$ on the equally discretized $x$-grid. 
The term $\mathbb{I}$ is a $\chi=1$ unit TT where all the element of its local tensors are $1$ as explained in the main text.
For $\vert\kappa\vert<1$ and $\vert w(x)\vert <1$, the iteration converges when starting from a unit-element TT $R^{(0)}=\mathbb{I}$ which represents $R^{(0)}(x) = 1$.
Then, the iteration converges rapidly due to the quadratic convergence of the Newton-Raphson method, such that the final TT, $\lim_{k\to\infty} \ket{R^{(k)}}$, accurately represents $(1 - \kappa w(x))^{-\frac{1}{2}}$.

\subsection{Conducting the self-consistent Picard iteration}

The Kidder equation is given by:
\[
w''(x) \;+\; 2\,x\,w'(x)\,\bigl(1 - \kappa\,w(x)\bigr)^{-\tfrac{1}{2}} \;=\; 0,
\]
subject to the boundary conditions
\[
w(0) \;=\; 1
\quad\text{and}\quad
\lim_{x\to\infty}w(x) \;=\; 0.
\]
To solve it numerically, we discretize the interval $[0,x_{\max}]$ into $2^n$ grid points $x_i$, with inter-grid point spacing $\Delta x= x_{\max}/2^n $. Here, $n$ represents the number of tensors required to store the discretized field in the TT format. 
For the Kidder equation with $\kappa=0.5$, 
it is sufficient to set $x_{\max}=16$ to approximate $x_{\max}\approx\infty$ \cite{parand2017new}, leading to 
the grid resolution of $\Delta x=2^{-12}$.
We label the approximate solution by $w_i \approx w(x_i)$ and use finite differences to approximate $w'(x)$ and $w''(x)$. 
For the interior points $i=1,\dots,2^n$, the discrete system typically reads as follows:
\begin{equation}
\begin{split}
F_i(\mathbf{w}) 
&\;=\;
\frac{w_{i+1} - 2\,w_i + w_{i-1}}{(\Delta x)^2}
\\
& ~~ \;+\;  2\,x_i \,\frac{w_{i+1} - w_{i-1}}{2\,\Delta x}\,\bigl(1-\kappa\,w_i\bigr)^{-\tfrac12}
\;=\; 0.
\end{split}
\end{equation}
We keep $w_0=1$ and $w_{2^n}=0$ fixed to enforce the boundary conditions.

A Picard fixed-point iteration solves this nonlinear system by rewriting each equation to isolate $w_i$ on one side. 
Concretely, one arranges:
\[
w_i 
\;=\; 
G_i\bigl(w_{i-1},\,w_i,\,w_{i+1}\bigr),
\]
so that each interior variable $w_i$ is expressed in terms of the old iteration values.  The explicit update formula, after algebraic rearrangement, reads
\begin{equation}
\begin{split}
G_i\bigl(\mathbf{w}^{(k)}\bigr) 
&=
\frac{w_{i+1}^{(k)} + w_{i-1}^{(k)}}{2} \\
&+
\frac{x_i\,\Delta x}{2}\,\Bigl[w_{i+1}^{(k)} - w_{i-1}^{(k)}\Bigr] \bigl(1-\kappa\,w_i^{(k)}\bigr)^{-\tfrac12}
,
\end{split}
\end{equation}
where $w_{i-1}^{(k)},\,w_{i+1}^{(k)}$, and $w_i^{(k)}$ come from the previous iteration. In TT representation, this Picard iteration becomes
\begin{equation}
\begin{split}
&G\bigl(w^{(k)}\bigr) = \frac{1}{2} \left( \hat{L} + \hat{R} \right) \cdot w^{(k)} \\
& + \frac{\Delta x}{2} X \odot \bigl(1-\kappa\,w^{(k)}\bigr)^{-\tfrac12} \odot \left( \hat{L} - \hat{R} \right) \cdot w^{(k)}.
\label{eq:LRops}
\end{split}. 
\end{equation}
Here, $X$ is a TT that encodes the coordinates of the discretized grid ($X^{[i]} = i\Delta x$)k,
$\hat{L}$ and $\hat{R}$ are linear operators that shift the index by one, 
i.e.~$\Psi' = \hat{L}\Psi$ s.t.~$\Psi'^{[j]} = \Psi^{[j+1]}$ and 
$\Psi'' = \hat{R}\Psi$ s.t.~$\Psi''^{[j]} = \Psi^{[j-1]}$ \cite{lubaschMultigridRenormalization2018}.
The TT representation of the term $(1 - \kappa w_i^{(k)})^{-\tfrac{1}{2}}$, 
denoted by $\bigl(1-\kappa\,w^{(k)\bigr)^{-\tfrac12}}$, 
is given by ITNTs in \eqnref{eq.kappaTransform}. 
We start from an initial guess $w^{(0)}$ and perform the iteration
\[
w^{(k+1)} = 
G\bigl(w^{(k)}\bigr),
\quad
k=0,1,2,\dots
\]
until successive solutions are sufficiently close. This fixed-point iteration converges to the unique TT $w$ 
satisfying the discretized Kidder equation and all of its boundary conditions.

\subsection{Anderson Acceleration}

While a basic Picard iteration is often sufficient for moderate grids or small parameter regimes, it can converge slowly or encounter numerical difficulties for larger or stiffer systems. Anderson Acceleration (also known as Anderson mixing) enhances the fixed-point iteration by forming and solving a small least-squares problem at each step, combining information from multiple past iterates and residuals to produce a more effective update.

Concretely, suppose we have at iteration $n$:
\[
w^{(k+1)} = G\bigl(w^{(k)}\bigr).
\]
Define the residual as
\[
f^{(k)} = G\bigl(w^{(k)}\bigr)  - w^{(k)}.
\]
Rather than simply setting 
\[
w^{(k+1)} = w^{(k)} + f^{(k)},
\]
we store several past pairs $\{w^{(k-i)},f^{(k-i)}\}$ and solve a small linear or least-squares system to find combination coefficients $\alpha_i$ that minimize the new residual. A commonly used variant enforces $\sum_i \alpha_i = 1$, preserving affine invariance. Then the next iterate can be written, for example, as
\[
w^{(k+1)}
\;=\;
\sum_{i} \alpha_i \,\bigl(w^{(k-i)} + f^{(k-i)}\bigr),
~\text{with}~
\sum_{i}\alpha_i=1.
\]
The small system to find a set of scalars $\{\alpha_i\}$ by computing dot products among the stored residuals. 
Once those $\alpha_i$ are determined, one combines the corresponding old solutions, 
$w^{(k-i)} + f^{(k-i)}=G(w^{(k-i)})$, to produce $w^{(k+1)}$. This linear combination step can drastically reduce the number of iterations required for convergence, especially in stiff or nonlinear regimes.

In practice, we limit the memory size (e.g., storing $m$ previous iterates) to keep the least-squares problem small. In a TT-based Kidder solver, the main computational effort lies in evaluating inner products of residuals $\mathbf{f}^{(k)}$ stored in tensor form and carefully truncating the TT rank to control memory usage. Anderson Acceleration thus provides a flexible, robust approach to speeding up Picard iteration for the discretized Kidder equation, particularly when direct methods (like Newton’s method) are more difficult to implement or too expensive at a large scale.

\subsection{Results}
We set the SVD tolerance to $10^{-12}$ with $\chi=400$, and conduct the Picard iteration for $10^5$ times with an Anderson acceleration with a memory of $m=10$. 
After this, we prolong the TT to a finer grid (see Ref.~\cite{lubaschMultigridRenormalization2018}) on which we re-run the above iteration. 
We start from a grid finesse of $N_{\text{res}}=4$ and end at $N_{\text{res}}=17$, the final result provides an initial slope of -1.1917912520439131, which shows an error of $\sim 6\times10^{-7}$ compared to the standard result given in Ref.~\cite{parand2017new}.

Here, we comment on the limitations of Picard iterations. To reach a high accuracy in the result (e.g., to find the initial slope to 10 digits), we need to improve the SVD tolerance to $\sim 10^{-14}$. 
In this case, each iteration step becomes slower as the bond dimension increases. Besides, as the grid finesse $N_{\text{res}}$ reaches $\sim18$, even the Anderson acceleration becomes slow to reach convergence; thus, we need a more efficient scheme to accelerate the Picard iteration, such as Newton's method. 
Neverthless, our ITNT enables the step of finding $(1-\kappa w(x))^{-0.5}$ via the Newton-Raphson method in the TT representation, 
and this approach is highly effective. 

\section{Truncation error bounding}\label{SM:trunceb}
For a positive field $\Psi$, with $ (\cdots)_{\mathrm{SVD}}$ denoting the field after SVD truncation, 
the maximum and the sub-maximum of $\Psi^{\odot 2^{p}}$ retain their order provided
\begin{align}
   \varepsilon (\Psi^{\odot {2^p}} , (\Psi^{\odot {2^p}})_{\mathrm{SVD}}) \leq 
   (\max \{ (\Psi^{\odot 2^p})^{[j]}  \} - \max_{\mathrm{sub}} \{ (\Psi^{\odot 2^p})^{[j]} \})/\sqrt{2}
\end{align}
as presented in \eqnref{eq:normreduction}.
In this appendix, we derive the right-hand side of this inequality by showing that 
$1/\sqrt{2}$ is the largest factor for which order preservation is guaranteed. 

Let $\Phi = \Psi^{\odot 2^{p}}$ and $\Phi' = (\Psi^{\odot 2^{p}})_{\mathrm{SVD}}$ for the exact and truncated arrays.
For simplicity
we write $\varepsilon = \left\| \Phi - \Phi' \right\|_2$ for the truncation error 
and $\Delta = \max\{ \Phi \} - \max_{\mathrm{sub}} \{ \Phi \}>0$ for the gap between maximum and the second maximal value 
of the untruncated field. 

Truncation displaces these two valus by 
\begin{align}
   \delta_1  = \max\{ \Phi \} - \max\{ \Phi' \}, \quad \delta_2 = \max_{\mathrm{sub}}\{ \Phi' \} - \max_{\mathrm{sub}}\{ \Phi \}
\end{align}
which measure the depression of the maximum ($\delta_1$) and the elevation of the second maximum ($\delta_2$).

In order for the order of the maximum and second maximum to be conserved, 
$\delta_1$ and $\delta_2$ must meet $\delta_1 + \delta_2 \leq \Delta$, while the truncation error irself constraints them by
$\delta_1^2 + \delta_2^2 \leq \varepsilon^2$.
Invoking Cauchy-Schwarz inequality, we obtain
\begin{align}
   \delta_1 + \delta_2 \leq \sqrt{2}\sqrt{ \delta_1^2 + \delta_2^2  } \leq \sqrt{2} \varepsilon \leq \Delta 
\end{align}
where the first inequality is saturated when $\delta_1 = \delta_2$, and the second when $\delta_1 = \delta_2=\varepsilon/\sqrt{2}$. 
With this, \eqnref{eq:normreduction} is shown. 

\section{Application of ITNT as Sign transformation - Accuracy and Scaling }\label{sec:sign_supMat}
To test the accuracy and the convergence properties of the sign transform, we consider the one-dimensional continuous function

\begin{figure*}[b!]
    \centering

    \begin{subfigure}[t]{0.27\linewidth}
        \centering
        \includegraphics[width=\linewidth]{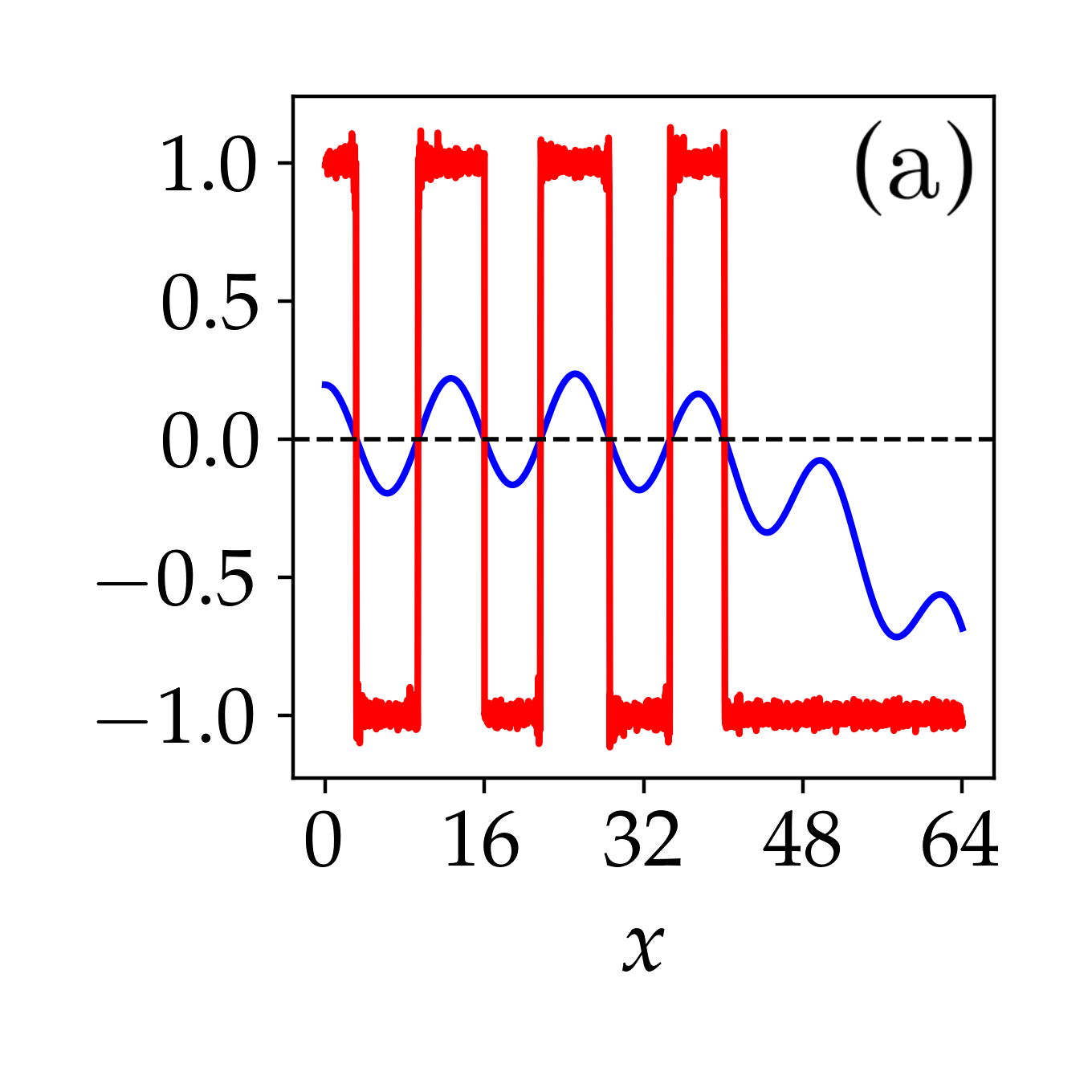}
        \label{fig:sign_N14_chi20}
    \end{subfigure}\hspace{-0.035\linewidth}
    \begin{subfigure}[t]{0.27\linewidth}
        \centering
        \includegraphics[width=\linewidth]{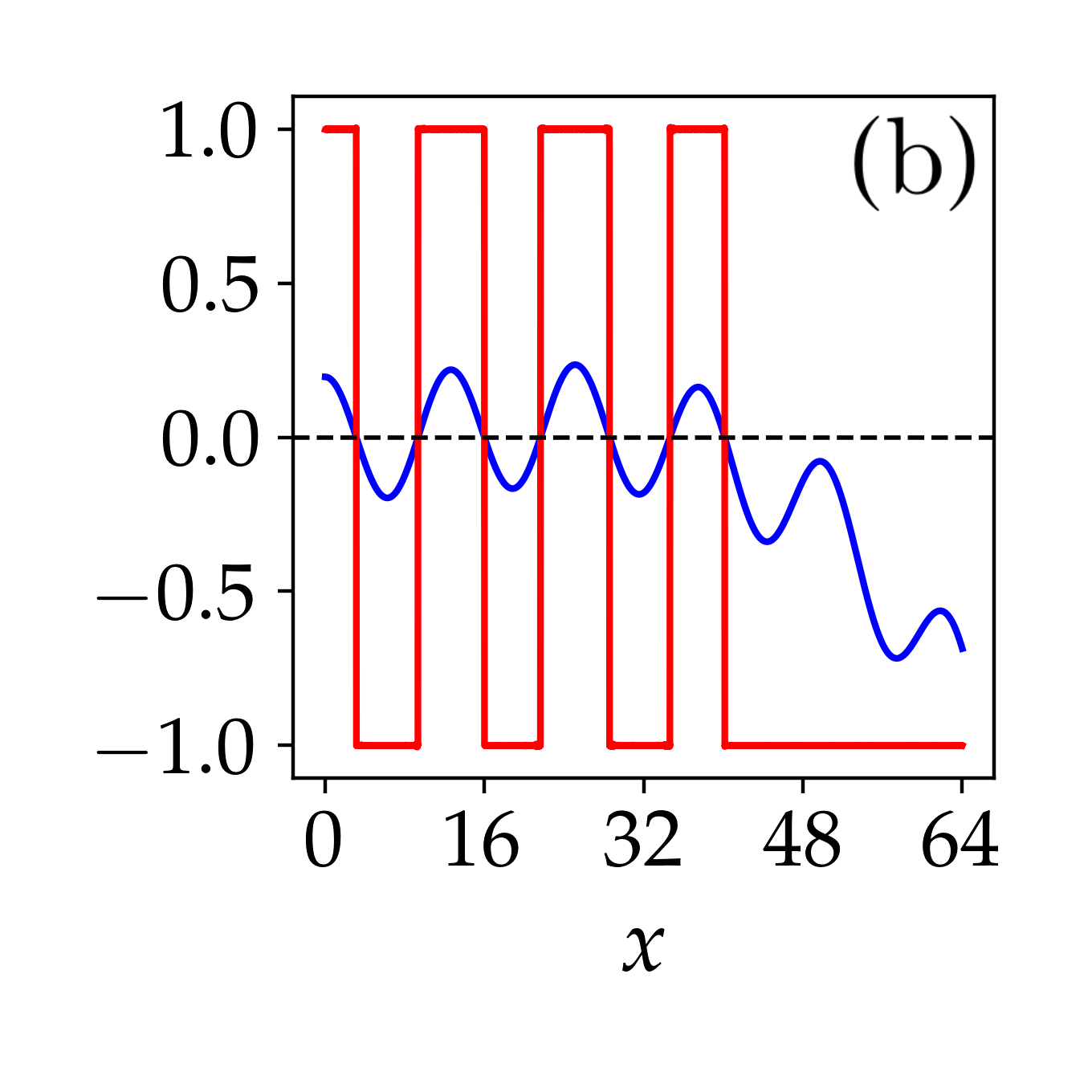}
        \label{fig:sign_N14_chi25}
    \end{subfigure}\hspace{-0.035\linewidth}
    \begin{subfigure}[t]{0.27\linewidth}
        \centering
        \includegraphics[width=\linewidth]{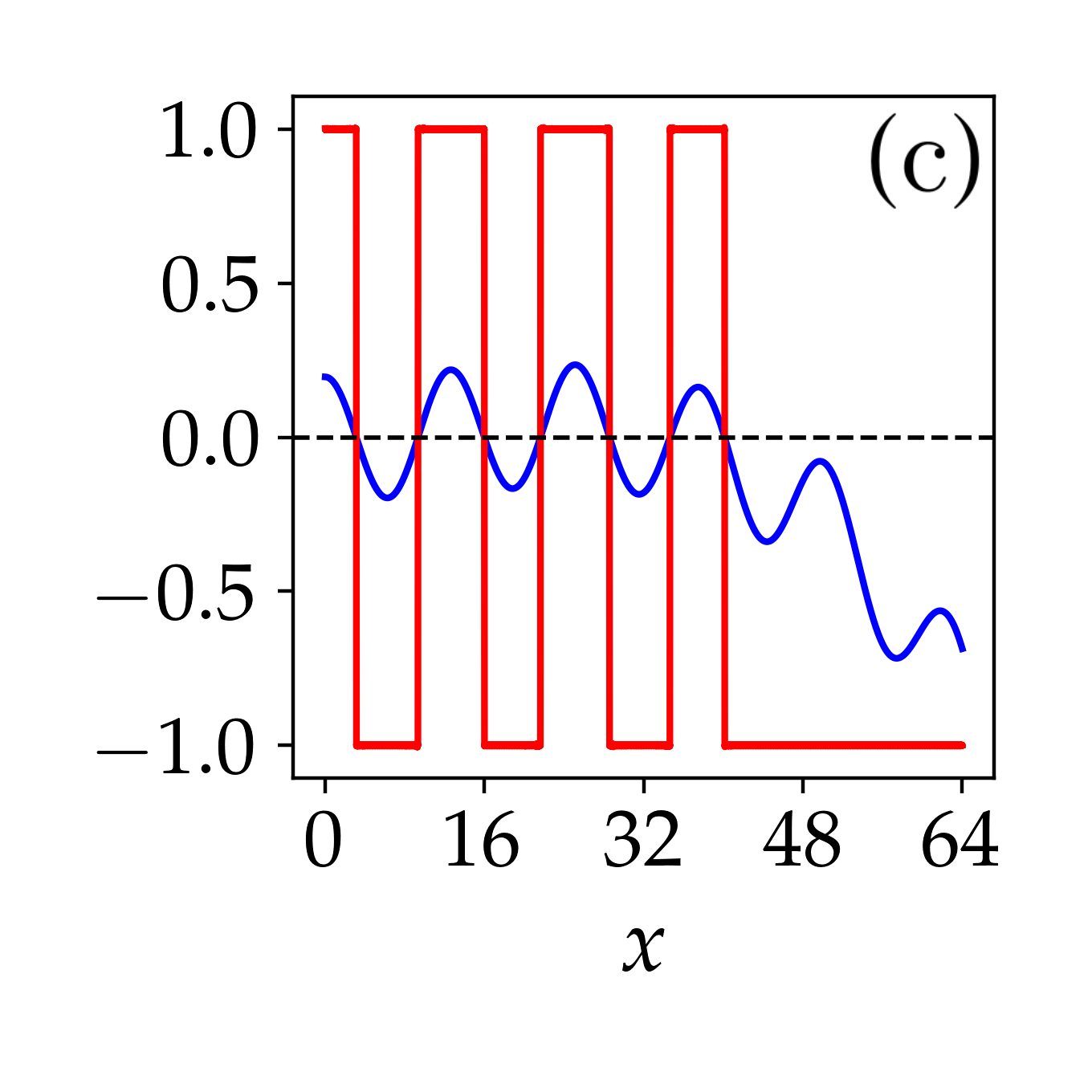}
        \label{fig:sign_N18_chi25}
    \end{subfigure}\hspace{-0.035\linewidth}
    \begin{subfigure}[t]{0.27\linewidth}
        \centering
        \includegraphics[width=\linewidth]{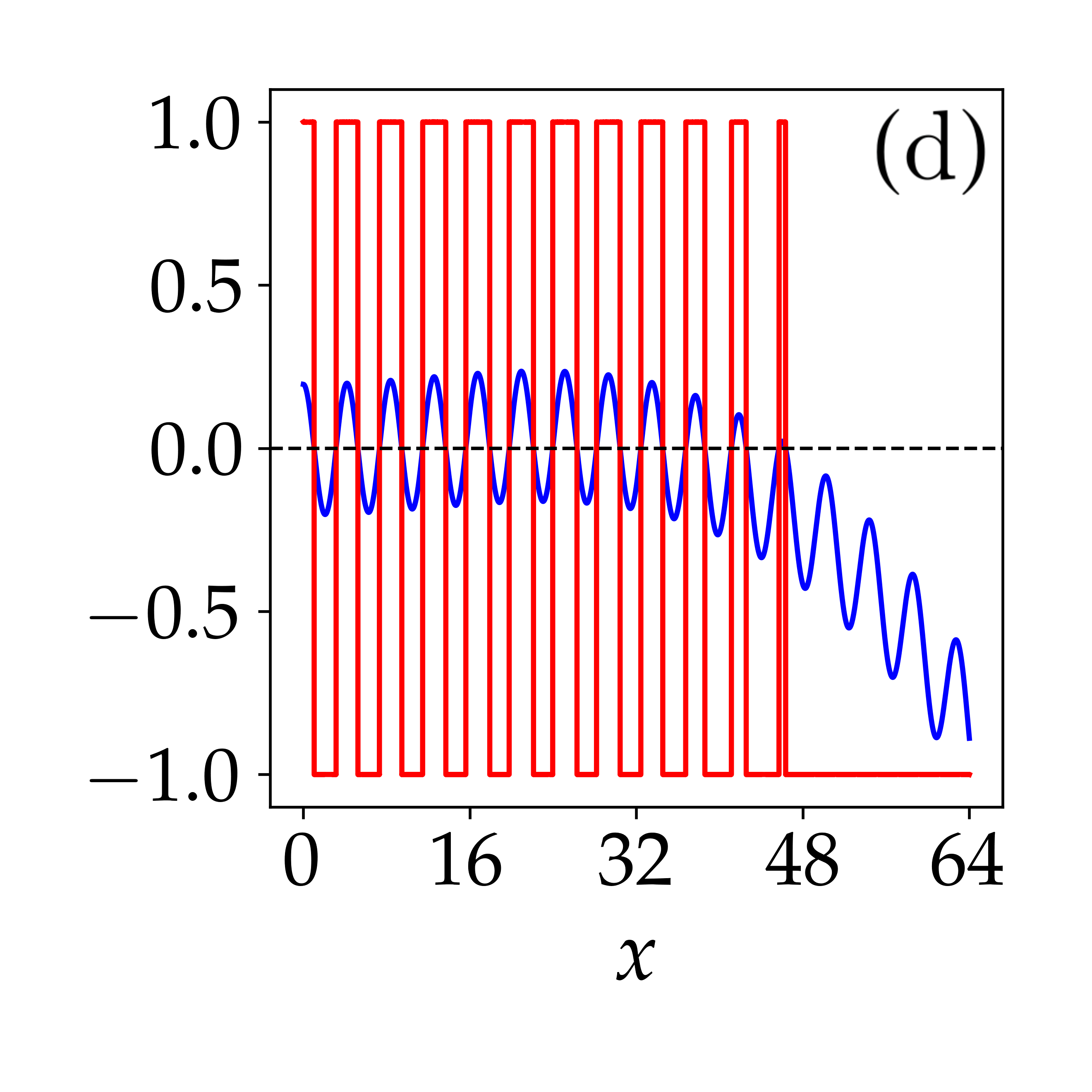}
        \label{fig:sign_N18_chi75_roots23}
    \end{subfigure}

    \vspace{-2.5em} %
    {\footnotesize
    \setlength{\tabcolsep}{0.5em} %
    \renewcommand{\arraystretch}{0.85} %
    \begin{tabular}{@{}c c@{}} %
        \raisebox{0.2ex}{\textcolor{blue}{\rule{1.2em}{1.4pt}}}\,$f_1(x)$ &
        \raisebox{0.2ex}{\textcolor{red}{\rule{1.2em}{1.4pt}}}\,$\mathrm{ITNT}(f_1(x))$
    \end{tabular}
    }
    \vspace{-0.6em} %

    \caption{Result of the Sign-function transformation via ITNT of the function defined in Eq.~\ref{eq:f1} for different number of roots \(N_{\text{root}}\) and different qubit numbers $n$. For all examples we consider \(N_{\text{iter}}=20\) and an SVD tolerance of \(10^{-8}\).
    Sign transformation of the function with the following properties:
    (a) \(n=14\), \(N_{\text{root}}=7\), \(\chi_{\text{max}}=20\).
    (b) \(n=14\), \(N_{\text{root}}=7\), \(\chi_{\text{max}}=25\).
    (c) \(n=18\), \(N_{\text{root}}=7\), \(\chi_{\text{max}}=25\).
    (d) \(n=18\), \(N_{\text{root}}=23\), \(\chi_{\text{max}}=75\). For \(N_{\text{root}}=23\), a noiseless sign transformation requires \(\chi_{\text{max}}=75\).}
    \label{fig:sign_transform_1x4}
\end{figure*}
\begin{figure*}[t!]
    \centering
    \includegraphics[scale=1.0]{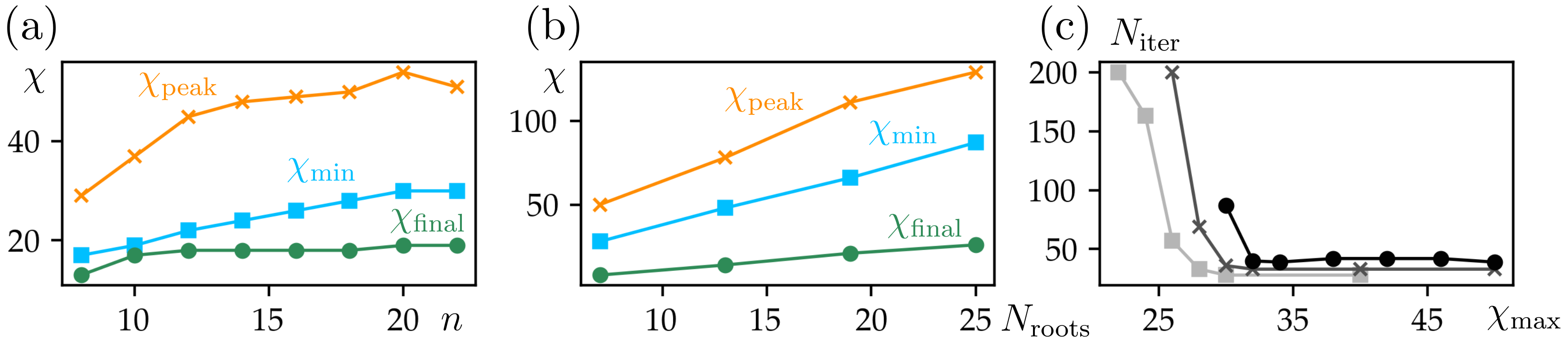}
    \caption{{\bf Scaling of the element-wise evaluation of sign function via ITNT}
       (a)~Element-wise evaluation of sign function $y=\sign\{f_\omega(x)\}$ (red) using ITNT
       of a function $y=f_\omega(x)$ (black) $\omega=1.0$. 
      (b)~Required bond dimension in dependence on the number of tensors $n$
      for $\omega=0.5$.
      Considered are the maximal intermediate bond-dimension $\chi_{\mathrm{peak}}$ (crosses) used by the algorithm if the bond dimension is not restricted and the SVD tolerance is fixed at $10^{-8}$. Furthermore, we consider the final bond dimension $\chi_{\mathrm{final}}$ (circles) required to represent $y$ accurate up to the order of the SVD tolerance of $10^{-8}$ with respect to the $L^2$-error. Finally, there is the cutoff bond dimension $\chi_{\mathrm{min}}$ (squares) which defines a lower cutoff bound,
      required to reach the result with ITNT up to this accuracy while reaching the bond dimension $\chi_{\mathrm{final}}$ within $200$ iteration steps.
      (c)~Required bond dimensions for different numbers of roots $N_{\mathrm{roots}}$ by using  
      $\omega = [0.5,1.0,1.5,2.0]$ corresponding to $N_{\mathrm{roots}} = [7,13,19,25]$
      and $n=18$ to converge within a maximum of $200$ iteration steps. The bond dimensions are defined as for (b).
      (d)~The number of iteration steps $N_{\mathrm{iter}}$ 
      needed to reach the final compressed result with respect to 
      the chosen maximal cutoff bond dimension $\chi_{\rm{max}}$ 
      for $\omega=0.5$ for $n=14,18,22$ (light to dark).
      All simulations were implemented using the zip-up algorithm for point-wise multiplication.}
       \label{fig:chi_and_steps_sign}
\end{figure*}

\begin{figure}[t!]
   \centering
   \includegraphics[width=0.5\linewidth]{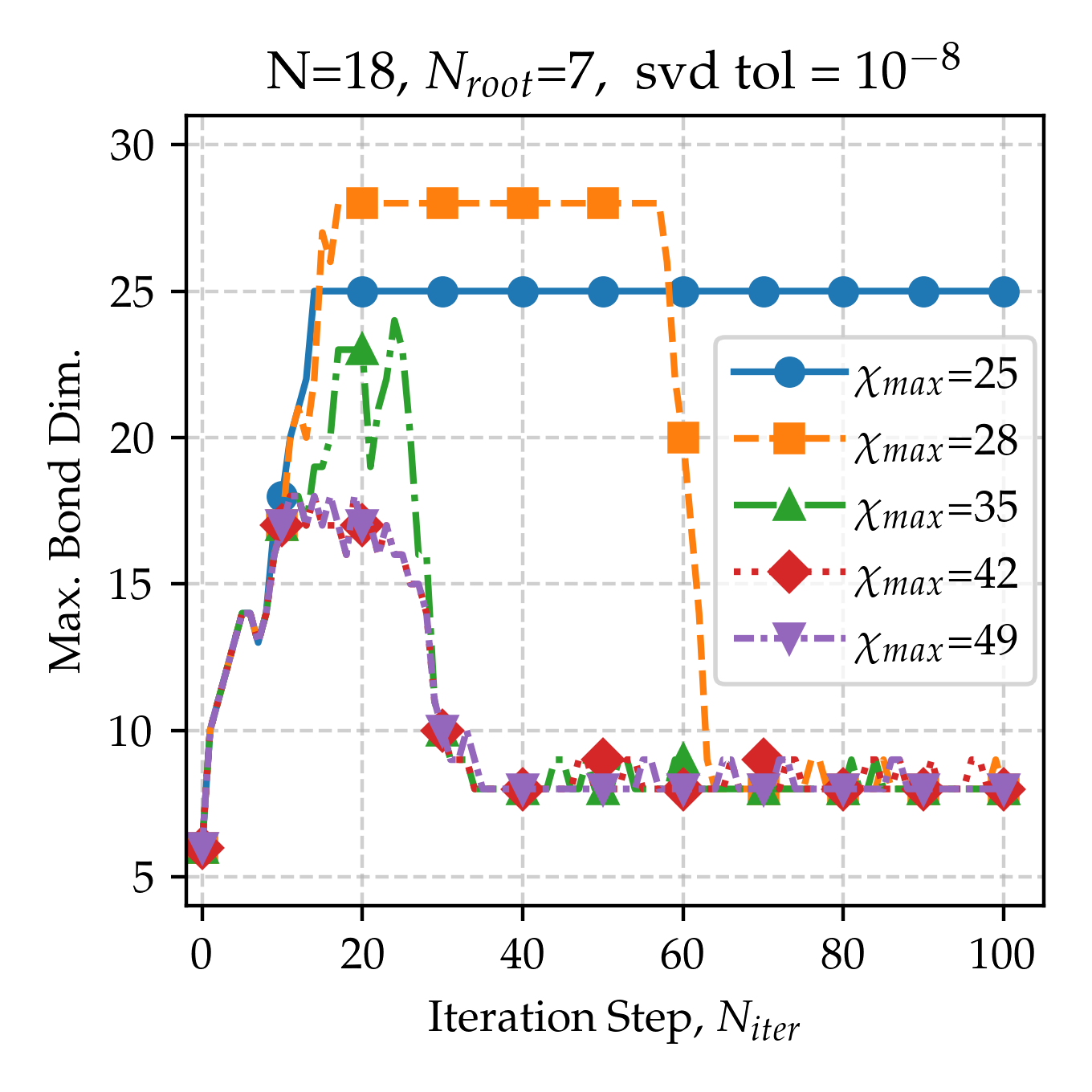}
   \caption{The bond dimension of the sign-transformed TT after $N_{\text{iter}}$ steps of iteration. 
   The qubit number is $n=18$ with root number $N_{\text{root}}=7$, and $\chi_{\text{max}}$ denotes the maximum bond dimension allowed during TT operations (e.g., element-wise products), with the SVD tolerance fixed at $10^{-8}$. For $\chi_{\text{max}} =28$, the convergence is reached at $N_{\text{iter}}\approx64$. For $\chi_{\text{max}} \gtrsim35$, the required iteration count reduces to $N_{\text{iter}}\approx34$.}
   \label{fig:MaxBondDim_vs_IterStep}
\end{figure}

\begin{equation} \label{eq:f1}
    f_\omega(x) = \cos(\omega x)  -0.000035(x-2)^3 + 0.001 x^2 -0.02
\end{equation}
encoded across $n$ tensors, where $x\in[0,64)$ and the frequency $\omega$ determines 
the number of roots on this interval.
Varying the number of roots, we study the impact of the chosen cutoff bond dimension $\chi_{\rm{max}}$ on the convergence of the algorithm.
The result of the ITNT-based sign-transformation of $f_\omega(x)$ is depicted in Fig.~\ref{fig:sign_transform_1x4} for different values of $\omega$, and hence different number of roots $N_{\rm{root}}$, and with a changing bond-dimension. 
For the simulation shown in Fig.~\ref{fig:sign_transform_1x4}~(a), the chosen cutoff bond dimension is visibly not large enough to ensure the full convergence of the simulation, leading to additional oscillations around the correct result.  
For the simulation results shown in  Fig.~\ref{fig:sign_transform_1x4}~(b)-(d), the chosen $\chi_{\rm{max}}$ is sufficient for convergence. Hereby, the necessary $\chi_{\rm{max}}$ seems to depend less on the number of grid points than on the number of roots.
A more detailed analysis of these dependencies is shown in Fig. \ref{fig:chi_and_steps_sign}~(a) and (b), where we study the required bond dimension over the number of tensors and number of roots.
Interestingly, the bond dimension required to represent the final result $\chi_{\rm{final}}$ up to an accuracy of order $10^{-8}$, saturates quickly with the number of tensors while showing a linear dependence on $N_{\rm{root}}$. 
Furthermore, we consider the minimum cutoff bond dimension $\chi_{\rm{min}}$ required during ITNT to allow convergence within $200$ steps. This $\chi_{\rm{min}}$ is significantly lower than the bond dimension $\chi_{\rm{peak}}$, 
the bond dimension used by the algorithm when no cutoff bond dimension is set, and only the SVD tolerance of $10^{-8}$ is defined. 
This indicates that choosing a cutoff bond dimension can be beneficial to bound the resource requirements. However, it can be beneficial to choose it slightly larger than $\chi_{\rm{min}}$ to allow for a quicker convergence (c.f. \ref{fig:chi_and_steps_sign}~(c)). 
In our simulations, we observe that selecting a sufficiently large bond dimension $\chi_{\text{max}} \gtrsim 5 N_{\text{root}}$ is crucial to minimize the convergent iteration step, $N_{\text{iter}}$.
Furthermore, we find that when the maximum bond dimension is large enough (i.e., $\chi_{\text{max}} \gtrsim 5 N_{\text{root}}$), the sign transformation ITNT fully converges after $N_{\text{iter}}\approx2N$ steps of iteration. For such a large enough $\chi_{\text{max}}$, after the ITNT reaches convergence, the bond dimension of the TT shrinks back roughly to the root number $N_{\text{root}}$. We can truncate the bond dimension to $\chi_{\text{max}}$ up to $\chi_{\text{max}} \sim 4 N_{\text{root}}$ (c.f.  \figref{fig:MaxBondDim_vs_IterStep}). 
In this case, the convergence can still be reached, although it is reached at a larger iteration step $N_{\text{iter}}$. As we truncate to an even smaller $\chi_{\text{max}}$, the iteration converges very slowly, and the whole iteration may eventually diverge due to numerical errors.

The runtime scalability with qubit number \(N\) is evident: as we increase \( N \), the bond dimension and runtime remain manageable. The runtimes required to perform the sign transformation with ITNT are summarized in Tab.~\ref{tab:runtimes_sign} and compared with a standard sign transformation using full-grid point-wise evaluation in Python/NumPy. Both computations were performed on the same laptop  with an Intel Core Ultra 9 275HX CPU.

\begin{table}[tb]
\centering
\begin{tabular}{|>{\raggedright\arraybackslash}p{1.5cm}|>{\raggedright\arraybackslash}p{1.5cm}|>{\raggedright\arraybackslash}p{1.5cm}|c|c|}
\hline
$n$ & $N_{\text{root}}$ & $\chi_{\text{max}}$ & time (sec) of ITNT-based sign tranforms & time (sec) of standard sign transform.  \\
\hline
22 & 7 & 45 & 3 & 1.6 \\
22 & 19 & 100 & 7 & 1.6 \\
22 & 51 & 250 & 65 & 1.6 \\
\hline
24 & 7 & 45 & 3.5 & 6.5 \\
24 & 19 & 100 & 8.5 & 6.5 \\
24 & 51 & 250 & 84 & 6.5 \\
\hline
26 & 7 & 45 & 5 & 28 \\
26 & 19 & 100 & 11 & 28 \\
26 & 51 & 250 & 110 & 28 \\
\hline
\end{tabular}
\caption{Comparison of runtime performance across different qubit numbers $n$,
root numbers $N_{\rm{root}}$,
and chosen cutoff bond dimensions $\chi_{\rm{max}}$.}
\label{tab:runtimes_sign}
\end{table}

A larger SVD tolerance can further boost the convergence speed, so we need to choose this tolerance properly. In our above simulations, we find that increasing the tolerance from $10^{-8}$ to $10^{-6}$ will only slightly increase the error of the ITNT result (i.e., the $L^2$-norm of the difference compared with the element-wise transformed array), from $\sim 10^{-8}$ to $\sim 10^{-6}$.

At the core of the sign-transformation lays the the homomorphic comparison iteration, $F(f) = 0.5 f (3 - f^2)$. In the following we will consider its iteration speed for different distances to the root. 
\paragraph{Near \( f = \pm 1 \): Double-exponential Convergence}

The stable fixed points of \(F\) are at \( f^* = \pm 1 \). Near \( f=1 \), let \( \epsilon^{(k)} = 1 - f^{(k)} \) represent the small deviation from the fixed point. The recurrence is approximately $\epsilon^{(k+1)} \approx 1.5\, (\epsilon^{(k)})^2$.
This leads to double-exponential convergence $\epsilon^{(k)} = (1.5\, \epsilon^{(0)})^{2^k}/{1.5}$.
If \(1.5 \epsilon^{(0)} < 1\), \(\epsilon^{(k)}\) decays extremely rapidly, ensuring that the ITNT quickly approaches \( f^* = 1 \). A similar argument applies near \( f=-1 \).

\paragraph{Near \( f=0 \): Exponential Departure and Resolution Threshold}

In contrast to the stable fixed points at \( f = \pm 1 \), the point \( f=0 \) is an unstable fixed point. To see this, we linearize \(F\) around \(f=0\):
\[
F(f) = 0.5 f (3 - f^2) \approx 1.5 f \quad \text{for small } f.
\]
Thus, any small nonzero value of $f$ near zero grows by approximately 1.5 per iteration, $f^{(k)} \approx f^{(0)} \times (1.5)^{k}$.
This results in exponential growth away from zero,  repelling from \( f=0 \). 
Suppose that after \( S \) steps of iteration, the iterative process realizes a sign-like function, i.e., \( f^{(S)} \approx 1 \), we can estimate the ``resolution threshold'' for the initial value, $f^{(0)} \approx (1.5)^{-S}$.

This threshold, \((1.5)^{-S}\), sets up a resolution limit in approximating the sign function. After \( S \) steps, any initial values \( f^{(0)} \) (and hence portions of the function \( f(x) \)) larger than \((1.5)^{-S}\) are effectively sign-transformed and pushed toward \( 1 \), while those smaller than \((1.5)^{-S}\) remain evidently far from 1. In other words, after sufficiently $S$ steps of iteration, the iterative map cannot distinguish initial value differences larger than \((1.5)^{-S}\) from 1, thus achieving a sign-like output up to that resolution level. 

Finally, we provide details of applying the ITNT-based sign function to locate the roots of smooth 1D functions. 
For a smooth function $f(x)$, the intermediate value theorem implies that whenever there is a change in the sign of $f(x)$, 
there exists at least one root of a function. 
When a curve is represented as an TT, roots can be found by first applying the sign-transformation ITNT, 
then, we apply the differential differential linear operator $\hat{R} - \hat{L}$ for the linear operators 
$\hat{R}$ and $\hat{L}$ introduced in \eqnref{eq:LRops} to obtain the TT that marks the root positions as delta peaks. 
From the resulting multi-delta-peak TT, we use the deflation method in Sec.~\ref{sec:INT_extremum-finding} to iteratively locate each peak position in the TT representation. 
This procedure outputs the set of root positions on the discretized grid.

\section{Benchmark: performance of extremum finding on discrete problem\label{SM:bench-2DNNI}}
Here, we focus on a classical Ising model on a square lattice of size $12 \times 12$
with nearest-neighbor interactions and on-site terms, 
where all parameters are drawn randomly from $(-0.5, 0.5)$ (c.f. Methods \ref{sec:IsingEncoding} for details of the model). The energy landscape of the classical Ising model can be mapped onto energy-landscape TT $E$ 
with the bond dimension that is proportional to the number of interaction terms in the Hamiltonian. 
For this example, with the open boundary condition, we found $\chi = 409$ achieves the exact encoding of the energy landscape. 
To identify the maximum energy configuration, we fix the Hamiltonian and perform 
self-multiplication $p=8$ times, then perform 800 independent deflation runs using 
100 different stochastic seeds. 

\begin{figure}[t!]
   \centering
   \includegraphics[width=0.99\textwidth]{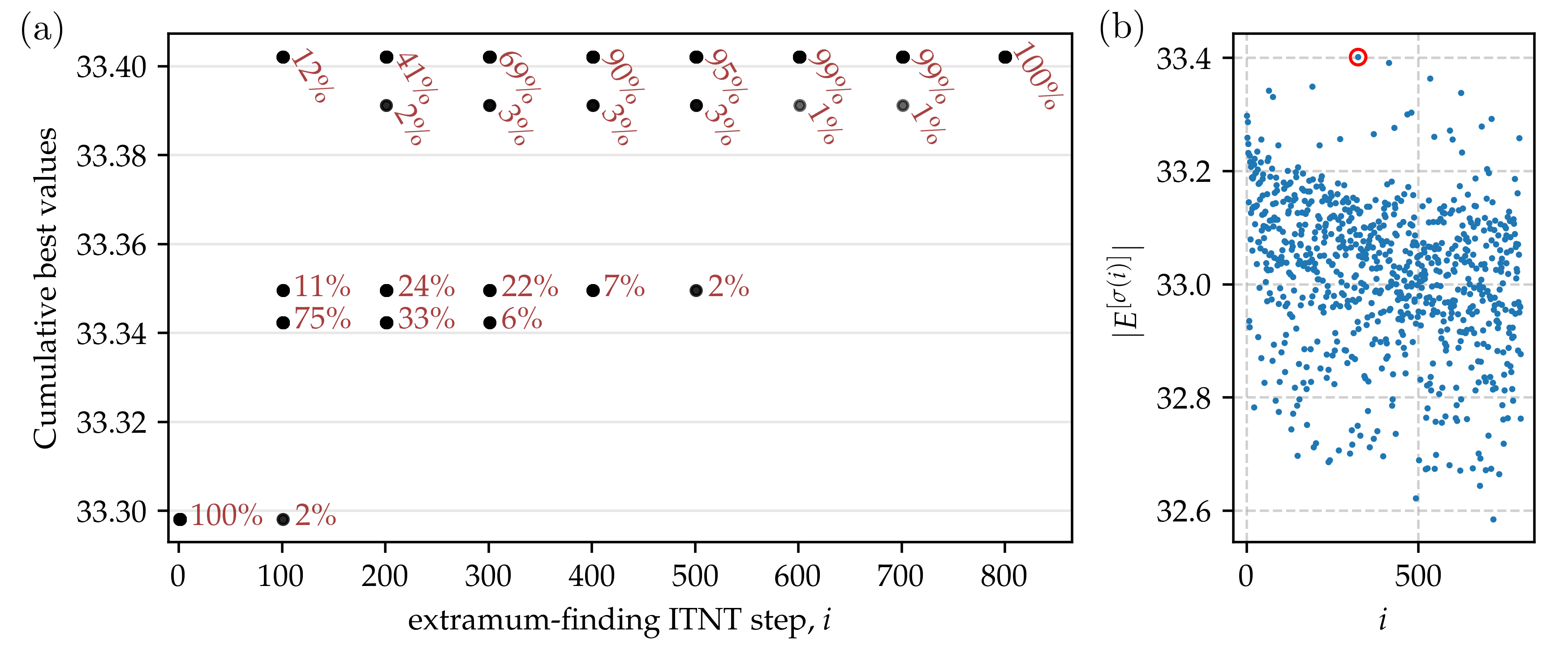}
   \caption{\textbf{Accuracy and scaling benchmarks.} 
      (a) Binary-discard deflation for locating extrema of a $12\times12$ classical Ising model. The algorithm is performed for 
      100 runs (each including 800 extremum-finding ITNT steps $i$), 
      and the cumulative best value at each step is shown, with the percentages indicating the fraction of runs that attain the corresponding value.
      (b) A single run of the extremum-finding ITNT steps. We fix $\chi=409$ and iteratively self-multiply $E$ 
      that encodes the energy of all the configurations 8 times. 
      In the resulting TT, we identified $800$ sub-leading peaks, 
      and found the extremum $\vert E\vert_{max}\sim33.35$ after the $i=324$\textsuperscript{th} step. 
      Here, the configuration returned by the $i$\textsuperscript{th} deflation step is denoted as 
      $\bm{\sigma}(i)$ with corresponding index $\sigma(i)$, and $E^{[\sigma(i)]}$ 
      is the corresponding energy of the found configuration. 
   }
   \label{fig:benchmark}
\end{figure}

In \figref{fig:benchmark}~(a), we present the statistics of the located extremum values 
with respect to the iteration steps for all 100 independent runs with the same Hamiltonian but different random site sequences for binary fixation.
The percentage on the right of each data point represents the fraction of runs 
that identified the value as the maximum energy configuration within those steps. 
While the subsequent ITNT steps identifies and converges to the true optimum,
as increasingly many trajectories find higher energy configurations over the iterations. 
In practice, 
each deflation step identifies a pronounced peak close to the optimal solution; 
after 800 stochastic deflation steps, 
all 100 runs successfully converge to the same maximum energy configuration. 
The detailed deflation process for a representative run is shown in \figref{fig:benchmark}~(b), illustrating how 
ITNT identifies the extremum within TT framework.

This task involves a 144-site system spanning a configuration space 
of $2^{144} \sim 10^{43}$. 
Performance-wise, a single self-multiplication of a $\chi = 409$ TT takes $\sim 2000$ 
seconds on an Intel Core Ultra 9 275HX CPU. 
which has been used for all computational benchmarks.
The subsequent deflation steps take between 3 and 30 seconds 
as the bond dimension grows from 409 to 1209. 
Finally, this extremum-location ITNT can be generalized to find configurations 
with energies close to arbitrary target values. 
This capability is discussed in detail in the following, 
where we sample the $E = 0$ configurations of the Ising model and find a configuration $\bm{\sigma} \in \{0,1\}^{n}$,
with the corresponding index $\sigma$, that satisfies $|E^{[\sigma]}|<10^{-4}$.

\subsection{Importance of self-multiplication steps}
\begin{figure}[t!]
    \centering
    \includegraphics[width=0.5\linewidth]{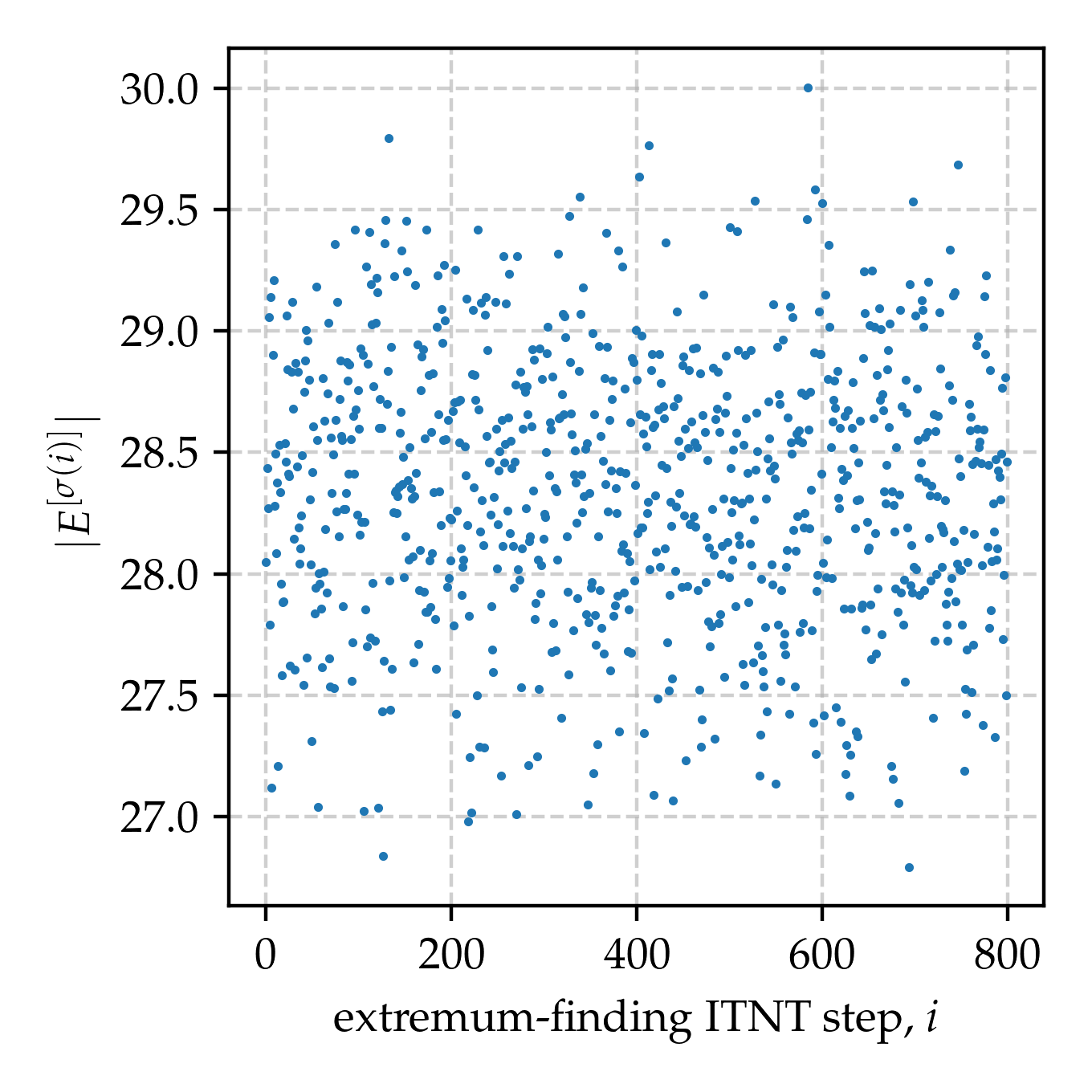}
    \caption{
   \textbf{Extremum-finding ITNT without preprocessing.}
   Shown is a single instance of the results obtained from 
   extremum-finding ITNT steps without performing the 
   self-multiplication preprocessing procedure.
   Clearly,
   in comparison to \figref{fig:benchmark}b.,
   the values of the found configurations are significantly lower,
   indicating a failure to detect the global extremum.
   }
    \label{fig:woselfmultiplication}
\end{figure}

To resolve tightly packed spectral features near the maximum in continuous data 
or degenerate configurations,
incorporating self-multiplication steps before the main 
extremum-finding ITNT routine is crucial.
In \figref{fig:woselfmultiplication},
we illustrate the configuration found throughout the 
extremum-finding ITNT steps.
As shown, without these self-multiplication preprocessing steps,
the algorithm fails to identify the correct extremum due to 
the model's rugged energy landscape.

\subsection{Application in energy matching}
\label{SM.sec.match_energy}
\begin{figure}[t!]
    \centering
    \includegraphics[width=0.5\linewidth]{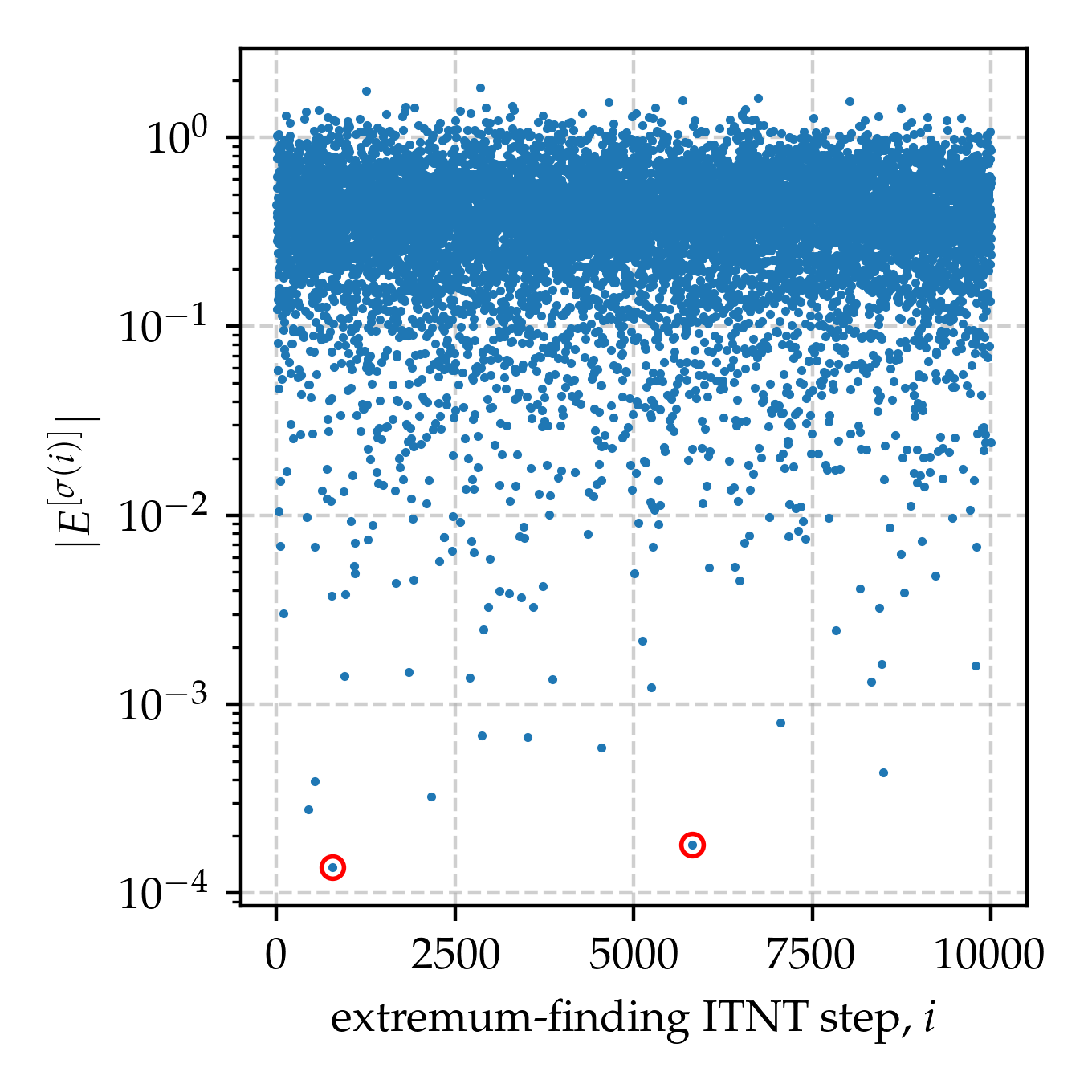}
    \caption{\textbf{Energy matching.} Energy matching $E({\bm \sigma})=0$ for the $144$-qubit model is conducted
       and 
       $10^4$ candidate are identified.
       The best match, $\vert E({\bm \sigma}_\text{best})\vert\sim1.38\times 10^{-4}$, is found after the $i=784$\textsuperscript{th} step. 
       The second-best match, $\vert E({\bm \sigma}_\text{next-best})\vert\sim1.81\times 10^{-4}$, 
       is found after the $i=5813$\textsuperscript{rd} deflation.
       During the self-multiplication, the bond dimension is fixed at $818$ for $14$ iterations (each taking $\sim 10^4$ seconds).
       During the following deflation, the bond dimension is truncated to $1900$ whenever it reaches $2000$, 
    yielding a per-extremum-finding ITNT runtime of $\sim60$ seconds.}
    \label{fig:energy_matching}
\end{figure}

In this section we detail the \emph{energy-matching} protocol referenced in the main text (Results), whose aim is to find configurations $\bm{\sigma}$ with $E(\bm{\sigma})\approx0$ for the same $12\times12$ random-field Ising instance. 
Following the notation from Methods~\ref{sec:IsingEncoding}), 
We encode the energy landscape as $E^{\sigma} = E(\bm{\sigma})$, and form the transformed landscape 
$E'(\bm{\sigma})=E(\bm{\sigma})^2-\lvert E\rvert_{\max}^2$ with 
$\lvert E\rvert_{\max}=\max_{\bm{\sigma}}\lvert E(\bm{\sigma})\rvert$ estimated via the extrema-locating ITNT. 
Since $E'(\bm{\sigma})\le0$, its extremum (most negative) value identifies the configuration with $E(\bm{\sigma})$ closest to zero, which we then locate using the same extremum-finding ITNT steps. 

The landscape of $E'({\bm \sigma})$ can be highly degenerate and rugged near its extrema. Thus, to locate the extremum of $E'({\bm \sigma})$, the deflation method is again very useful to mitigate the error arising from the bond-dimension truncation, allowing us to find a string matching $E({\bm \sigma})$ closer to $0$. 
We find that after performing the iterative self-multiplication for 14 times 
(with bond dimension fixed at a larger value $\chi=818$ to ensure a better accuracy of the ITNT algorithms), 
the random binary fixation finds matching accuracy at $E({\bm \sigma}_\text{first-try})\sim 10^{-1}$, 
while the subsequent iterative deflation (locating $10^4$ sub-leading peaks) improves the matching accuracy 
to $E({\bm \sigma}_\text{best})\sim 1.38\times10^{-4}$. 
Notably, in \figref{fig:energy_matching}, the best matching configuration separates from the next-best configuration, $ E({\bm \sigma}_\text{next-best})\sim 1.81\times10^{-4}$, by a Hamming distance of $45$. 
This rugged, multi-modal landscape severely challenges both gradient descent and sampling-based methods such as Parallel Tempering, which often get trapped in local minima or fail to access degenerate solutions, where our TT deflation method inherently bypasses.

\end{document}